\documentclass[11pt,a4paper]{article}

\usepackage{times,latexsym}
\usepackage{url}
\usepackage[T1]{fontenc}

\usepackage{ACL2023}

\usepackage{graphicx}
\usepackage{amsmath}
\usepackage{amssymb}
\usepackage{times}
\usepackage{latexsym}
\usepackage{tikz}
\usepackage[edges]{forest}
\usepackage{booktabs}
\usepackage{multicol}
\usepackage{subcaption} 

\usepackage{flexisym}
\usepackage[T1]{fontenc}

\usepackage[utf8]{inputenc}

\usepackage{microtype}

\usepackage{inconsolata}

\usepackage{siunitx}
\usepackage{multirow}
\usepackage{makecell}

\usepackage{comment}
\usepackage{rotating}
\usepackage{float}
\usepackage{colortbl}
\usepackage{graphicx}

\usepackage{xurl}
\usepackage{hyperref}

\usepackage{xspace,mfirstuc,tabulary}

\definecolor{mydarkblue}{RGB}{0,70,140} 

\title{Beyond the Explicit: A Bilingual Dataset for Dehumanization Detection in Social Media}

\author{
Dennis Assenmacher\textsuperscript{1\thanks{~~Corresponding author. Email: \texttt{first.last@gesis.org}}},
Paloma Piot\textsuperscript{2}, 
Katarina Laken\textsuperscript{3}, 
David Jurgens\textsuperscript{4} \and
Claudia Wagner\textsuperscript{1,5}\\
\textsuperscript{1}GESIS - Leibniz Institute for the Social Sciences \\ 
\textsuperscript{2}IRLab, CITIC Research Centre, Universidade da Coruña \\  
\textsuperscript{3}Fondazione Bruno Kessler \& Universidade de Santiago de Compostela \\
\textsuperscript{4}University of Michigan \\
\textsuperscript{5}RWTH Aachen University 
}

\begin{document}
\maketitle

\begin{abstract}

Digital dehumanization, although a critical issue, remains largely overlooked within the field of computational linguistics and Natural Language Processing. The prevailing approach in current research concentrating primarily on a single aspect of dehumanization that identifies overtly negative statements as its core marker. This focus, while crucial for understanding harmful online communications, inadequately addresses the broader spectrum of dehumanization. Specifically, it overlooks the subtler forms of dehumanization that, despite not being overtly offensive, still perpetuate harmful biases against marginalized groups in online interactions. These subtler forms can insidiously reinforce negative stereotypes and biases without explicit offensiveness, making them harder to detect yet equally damaging. Recognizing this gap,
we use different sampling methods to collect a theory-informed bilingual dataset from Twitter and Reddit. Using crowdworkers and experts to annotate 16,000 instances on a document- and span-level, we show that our dataset covers the different dimensions of dehumanization. This dataset serves as both a training resource for machine learning models and a benchmark for evaluating future dehumanization detection techniques. To demonstrate its effectiveness, we fine-tune ML models on this dataset, achieving performance that surpasses state-of-the-art models in zero and few-shot in-context settings.


\color{red}{This work contains harmful language}
\end{abstract}

\section{Introduction}
Dehumanization is a psychological process in which individuals or groups are described or treated as less than fully human, often by likening them to animals, objects, or machines. This process de-emphasizes the humanity of the targeted group either partially or entirely, thereby weakening moral obligations toward them \cite{haslam-review}. As a result, dehumanization can lead to extreme intergroup bias, heightened anxiety, hate speech, and violence directed at specific social groups. It often facilitates moral exclusion, where certain populations are perceived as existing outside the boundaries of moral concern. Such perceptions have historically enabled atrocities, war crimes, and genocides. Dehumanizing rhetoric remains prevalent in contemporary discourse. A recent statement by the current president of the United States illustrates this: \textit{“Democrats said, ‘Please don’t call immigrants animals.’ I said, no, they’re not humans, they’re animals”} \cite{laynenathanTrumpCallsMigrants2024}. This example shows how explicitly dehumanizing language continues to shape public narratives today \cite{khan2024israel, smithgaler2023dehumanising}.

To better understand the prevalence of digital dehumanization as well as strategies for countering it, the development of specialized detection methods is crucial. Our literature review shows that only a few computational approaches and datasets explicitly target the phenomenon of dehumanization. Existing work primarily focuses on a single aspect of dehumanization, often identifying overtly offensive and direct animalistic comparisons as the core marker. As a result, more subtle forms such as mechanistic dehumanization or indirect expressions are frequently overlooked. Furthermore, existing resources are almost exclusively limited to the English language. General-purpose toxicity or hate speech classifiers are also not well suited to detect dehumanization, since they are typically optimized to identify explicit abuse or profanity. They tend to miss language that denies human qualities without relying on overtly offensive expressions. Dehumanization requires task-specific modeling because it often manifests in implicit, metaphorical, or morally exclusionary language that may not be flagged by standard classifiers. This work addresses these limitations by introducing a large bilingual dataset of more than 490,000 dehumanization candidates, capturing multiple dimensions of dehumanization across two social media platforms: $\mathbb{X}$ (formerly Twitter) and Reddit. In addition, we conduct a crowd-sourced and expert-based annotation study in which at least three annotators produced document- and span-level annotations for in total 16,000 candidates.

Our comparison of 14 different methods on the binary dehumanization detection for each sub-dimension of the construct shows that all models trained on our dataset show substantial performance gains. Even in the few-shot setting, where only a small number of examples are provided in the prompt, performance improved by an average of 12\% compared to the corresponding zero-shot scenarios. Our cross-lingual analysis also demonstrates that our annotated data enables robust model performance even in cross-lingual applications.



\section{Background}
Conceptualizing dehumanization presents a significant challenge due to the lack of a universally accepted definition, both in psychology and in computational linguistics or natural language processing. One major difficulty lies in the fact that dehumanization is not always expressed explicitly or with overt negativity. For example, \citet{Loughnan2007} found that certain social groups are implicitly associated with non-human traits or beings even in the absence of direct hostility or negative evaluation. This highlights that dehumanization can operate subtly, without relying on clearly abusive or offensive language. The implication is that not all instances of dehumanizing language are easily recognizable as such, which complicates both conceptual clarity and computational detection. As a result, identifying dehumanization requires attention not only to explicit insults or comparisons but also to more implicit forms that reduce perceived humanity without overt aggression.

Despite the complexity of the concept, operational definitions currently employed in computational work predominantly focus on explicit forms of dehumanization \cite{engelmann-etal-2024-dataset, kirk-etal-2023-semeval, vidgen-etal-2020-detecting, vidgen-etal-2021-learning}. These definitions often center on overt comparisons to animals or monsters and neglect more implicit expressions that lack direct hostility but nonetheless reduce perceived humanity. 
In this section, we review existing theories and operationalizations of dehumanization in NLP and computational linguistics to identify gaps in how the construct is currently measured and modeled.
\subsection{Theories of Dehumanization}
Dehumanization has been extensively studied in psychology, leading to the development of various theories \footnote{A thorough report of existing Dehumanization Theories can be found in \citet{haslamDehumanizationInfrahumanization2014}}. \citet{Bar-Tal1989} introduces dehumanization as a means of delegitimization. According to him, dehumanization involves categorizing a group as ``subhuman creatures such as an inferior race or animals, or by using categories of negatively evaluated superhuman creatures such as demons, monsters, and satans'' \cite{Bar-Tal1989}. \citet{leyens_infra} introduced the notion of infrahumanization to capture the more nuanced and widespread practice of withholding secondary emotions from outgroup members, thereby differentiating it from the broader concept of dehumanization.
The Dual Model of Dehumanization, proposed by \citet{haslam-review}, distinguishes between two forms: animalistic and mechanistic dehumanization. Animalistic dehumanization involves the denial of attributes that are unique to humans, such as intelligence, civility, rationality, or morality. In contrast, mechanistic dehumanization involves the denial of attributes that are inherent to human nature, including emotional responsiveness, warmth, and agency.

Complementing these perspectives, the Stereotype Content Model (SCM), articulated by \citet{fiske_scm}, posits that our perceptions of others are framed along two primary dimensions: warmth and competence. Groups perceived as low in warmth and low in competence may elicit envious or paternalistic prejudices, which can include elements of dehumanization.
This theory, however, does not proceed from an explicit definition of humanness.
Therefore, we build our work on the Dual Model of Dehumanization, proposed by \citet{haslam-review}.

\subsection{Measuring Dehumanization: Computational Methods and Datasets}
\label{sec:computational}

\paragraph{Methods}
The work by \citet* {mendelsohn} proposes a  framework to detect dehumanization that primarily focuses on lexical indicators of dehumanization. Also the work by \citet* {landry2022} relies on textual indicators using the Mind Perception Dictionary \cite{schweitzer2021} which enables to study the prevalence of experience- and agency-related words. \citet{card2022} extended upon \citet* {mendelsohn} and developed a method that is based solely on contextual sentence information. By masking the target identity terms in a sentence they used a BERT model to predict the likelihood of the MASK token to be replaced by metaphorical terms representing dehumanization (e.g. different animal terms). The overall dehumanization score is calculated as the sum of the probabilities for all metaphors. \citet{engelmann-etal-2024-dataset} fine-tuned different HateBERT-based models on unlabeled data that was assigned a binary label according to the presence of keywords related to different types of dehumanization. 
\citet{zhang2024hatespeechnlpschallenges} use LLMs to identify dehumanizing language in the \citet{vidgen-etal-2021-learning} dataset. 
\citet{friedman2021toward} identify dehumanization using a theory-driven schema based on knowledge graphs, which capture entities, attributes (e.g., dehumanized), and their relations. To classify texts into this schema, they use SBERT embeddings on a dataset of 378 annotated posts from five online communities. In a recent work, \citet{mendelsohn2025peoplefloodsanalyzingdehumanizing} developed a computational approach to detect dehumanizing metaphors in immigration discourse on social media. They identified seven source domains commonly evoked in such discourse (e.g., water, vermin). To detect metaphorical language associated with these domains, they employed LLMs to identify metaphorical words and utilized document embeddings to capture metaphorical associations in context. This method does not rely on manual annotation but instead uses brief concept descriptions and a handful of example sentences for each metaphorical concept. 

Most prior work focused on English language only and/or on a binary concept of dehumanization directed at specific target groups (e.g., Ukrainians \cite{burovova-romanyshyn-2024-computational}, women \citet{kirk-etal-2023-semeval}). Our work aims to broaden the scope by examining how state of the art language models can be adapted to detect different theoretical dimensions of dehumanization in two different languages and a wider range of target groups discussed in the literature.

\paragraph{Datasets}

Supervised or semi-supervised machine learning methods require annotated data to train and test models that measure different aspects of dehumanization. 
Recently \citet{burovova-romanyshyn-2024-computational} presented a computational analysis of dehumanization in the context of extreme violence, focusing on dehumanizing rhetoric toward Ukrainians on Russian social media, particularly Telegram. They annotated 4,111 sentences for binary dehumanization classification and 478 sentences for span-level dehumanization types, differentiating between denial of human uniqueness and denial of human nature.  
\citet{engelmann-etal-2024-dataset} released an English-language dataset of dehumanization on Reddit and political forums from the CommonCrawl corpus. They published an annotated dataset consisting of 916 instances. 
\citet{sachdeva-etal-2022-measuring} introduce the Measuring Hate Speech (MHS) corpus, a dataset developed to assess hate speech while incorporating annotators’ perspectives through Rasch Measurement Theory. Dehumanization is included as one of the survey items used to judge the hatefulness of comments, enabling annotators to flag content that dehumanizes targeted groups. The dataset primarily captures explicit, animalistic forms of dehumanization and contains 2,476 clear instances, defined as comments with an average dehumanization rating of 3 or higher on a 0–4 scale.
\citet{vidgen-etal-2021-learning} propose a new dynamic process for improving online hate detection through a human-and-model-in-the-loop approach, resulting in the creation of a dataset with approximately 40,000 entries, including about 15,000 challenging perturbations. This novel dataset, which includes fine-grained labels (one of them is Dehumanization) for the type and target of hate, significantly enhances model performance on hate detection tasks. 
Their dataset consists of 962 dehumanization instances. \citet{kirk-etal-2023-semeval} publish a dataset on online sexism. In one of their finer-grained categories they capture explicit forms of dehumanization towards women. 
Their dataset consists of 286 dehumanizing instances from Twitter and Gab.
\citet{animal_metaphor} identifies dehumanizing animal metaphors used for people fleeing Wuhan, in the context of the outbreak of the Coronavirus, in December 2020. In a related study, \citet{vidgen-etal-2020-detecting} looks at abuse towards East Asian people, with Dehumanization being one of the derogatory categories. They also focus on the explicit animalistic dehumanization dimension, mainly searching for comparisons between East Asians and non-humans or subhumans, such as insects, weeds, or actual viruses. They identify 104 instances that reveal dehumanizing signals.
\citet{joshi2024} introduced a dataset consisting of 10,638 instances, focusing exclusively on animalistic dehumanization. They began by compiling a list of 75 terms commonly used to compare people to animals and collected Reddit posts containing these terms. To curate the dataset, they used Flan-T5 XXL to sample candidates, aiming for a distribution of 70\% likely positive and 30\% likely negative instances. After annotation, 6,884 examples were labeled as dehumanizing.

Unlike prior dataset, our dataset covers different theoretical dimensions of dehumanization (i.e., goes beyond animalistic dehumanization), offers document- and span-level annotations, stretches across two different platforms, different targets and languages and spans over ten years.

\section{Operationalizing the Dual Model of Dehumanization}
Our taxonomy is inspired by the conceptual grid for denials of humanness proposed by \citet{bain-haslam-book}. Acknowledging that there is no \textit{``crisp definition of dehumanization waiting to be discovered''}, he developed an abstract two-dimensional model with the intention of providing a frame for multiple taxonomies. 
\begin{figure*}[ht!]
    \centering
    \resizebox{0.95\textwidth}{!}{%
    \begin{forest}
        for tree={
            align=center,
            font=\small,
        },
        forked edges, 
        [Dehumanization
            [Animalistic
                [Explicit
                    [{Comparing one of the targets  \\with an \emph{animal} or \emph{subhuman}  \\(e.g. cockroaches or diabolo)}]
                ]
                [Implicit
                    [{Describing one of the targets \\as being incivil, coarse, irrational, \\ child like or lacking of moral}]
                ]
            ]
            [Mechanistic
                [Explicit
                    [{Comparing one of the \\targets with a machine \\ (e.g. robot, automaton)} ]
                ]
                [Implicit
                    [{Describing one of the \\targets as cold, inert, rigid, \\ superficial or lacking of agency}]
                ]
            ]
        ]
    \end{forest}
    }
    \caption{Conceptualization of Dehumanization based on \cite{haslam-review,bain-haslam-book}}
    \label{fig:smart_card_types}
\end{figure*}
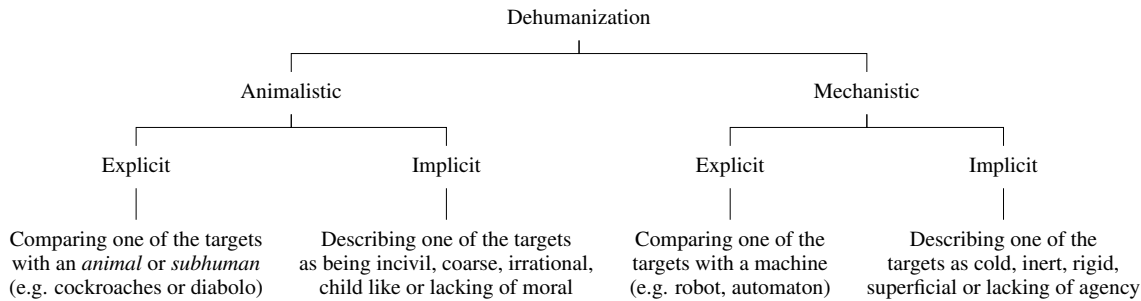

\subsection{Animalistic / Mechanistic}

\paragraph{Animalistic Dehumanization:} This form of dehumanization occurs when people are thought of as animals or subhumans, thereby stripping them of uniquely human characteristics (UH) such as rationality, morality, and civility. This type of dehumanization is often applied to ethnic and racial outgroups and has been the subject of significant scholarly focus. For example, there have been instances where certain racial groups have been likened to apes or insects.
\paragraph{Mechanistic Dehumanization:} This involves the denial of human nature (HN), viewing people as machines, lacking traits such as emotionality, warmth, cognitive openness, individual agency, and depth. While these traits are central to what it means to be human, their absence doesn't make someone more animal-like but rather machine or automaton-like. According to \citet{haslam-review}, inertness, coldness, rigidity, fungibility, lack of agency and empathy. Compared to animalistic dehumanization, mechanistic dehumanization has been less studied and its role in intergroup conflict is less clear. It has been most examined in the context of the objectification of women, where people are reduced to mere instruments for the use of others \citep{doi:10.1177/0146167212436401}.

\subsection{Explicit / Implicit Dimension}

\paragraph{Explicit Dehumanization:} This is characterized by more direct, often conscious, denials of humanness to a target. This can involve judgments that refer directly to nonhuman entities, like animals, apes, or robots. Research in this area often looks at explicit associations between groups and nonhuman entities. This form of dehumanization is more overt and clear-cut compared to subtle dehumanization. It should be emphasized that not all explicit and direct comparisons to animals or machines are necessarily dehumanizing in a sense that they causing harm to people \cite{over}. \textit{He is strong like a lion} or referring to an athlete as a \textit{machine} are direct animalistic/mechanistic comparisons that has a positive connotation. 
\paragraph{Subtle Dehumanization:} This type of dehumanization encompasses phenomena that are more understated and can occur in everyday situations, without overt intergroup antagonism or conflict and independent of any negative evaluation of the other group. 
Infrahumanization is a key example of subtle dehumanization \cite{leyens-infra}. It was conceptualized as a subtle phenomenon that can occur in the absence of overt conflict and is not directly referred to as ``dehumanization'' by researchers in an effort to distinguish it from more extreme forms \cite{bain-haslam-book}. According to \citet{haslam-review} these are in its animalistic form the denial of UH and in its mechanistic form denial of UN traits.

\section{The Dataset(s)}
\label{sec:thedataset}

The dataset creation process involves information-retrieval and preprocessing steps, including a theory-driven approach to identifying potentially dehumanizing data instances. In the following, we describe the overall process and the data sources in more detail \footnote{Pseudonymized candidate text data and annotated instances (MASKED users and urls) will be shared with researchers upon request}.

\subsection{Data Sources}
The datasets utilized in this study originated from a comprehensive $\mathbb{X}$ dataset, formerly known as Twitter, collected over a span of 10 years (from 2013 to 2023) \cite{kts-twitter}. This dataset comprises 1\% of all tweets, encompassing data from January 2013 onwards, obtained through the public Twitter API. Furthermore, this Twitter data was enriched by four years (2019-2022) of Decahose content (10\% data stream). We also draw on Reddit data collected via the Pushshift API over the same period (2013–2023) \cite{baumgartner2020pushshift}. The Pushshift dataset includes Reddit posts, comments, and associated metadata; however, for this study, we focus exclusively on user-generated comments. We restrict our study to tweets and Reddit comments posted in English or German. Also we only consider original content, i.e. no quotes or retweets.

\subsection{Candidate Selection} 

Our study investigates a broad spectrum of potentially dehumanizing messages. To manage the large volume of data, we initiated the analysis by filtering messages that reference possible targets of dehumanization. This included explicit mentions of specific groups, as well as messages using personal pronouns like “they” or demonstrative pronouns like “those” and “these” to imply group references.

The list of target groups is based on prior research identifying populations particularly vulnerable to dehumanization \cite{hodson2007interpersonal, goff2008not, vaes2011, loughnan2014}, and was expanded through manual inspection and semantic similarity using fastText embeddings trained on our dataset. This allowed us to capture direct mentions, variations, and colloquial forms, strengthening our filtering approach. The final lists of target terms for both languages are provided in Appendix Tables~\ref{tab:terms},\ref{tab:terms_german}.

\begin{figure*}
    \centering
    \includegraphics[width=0.9\textwidth]{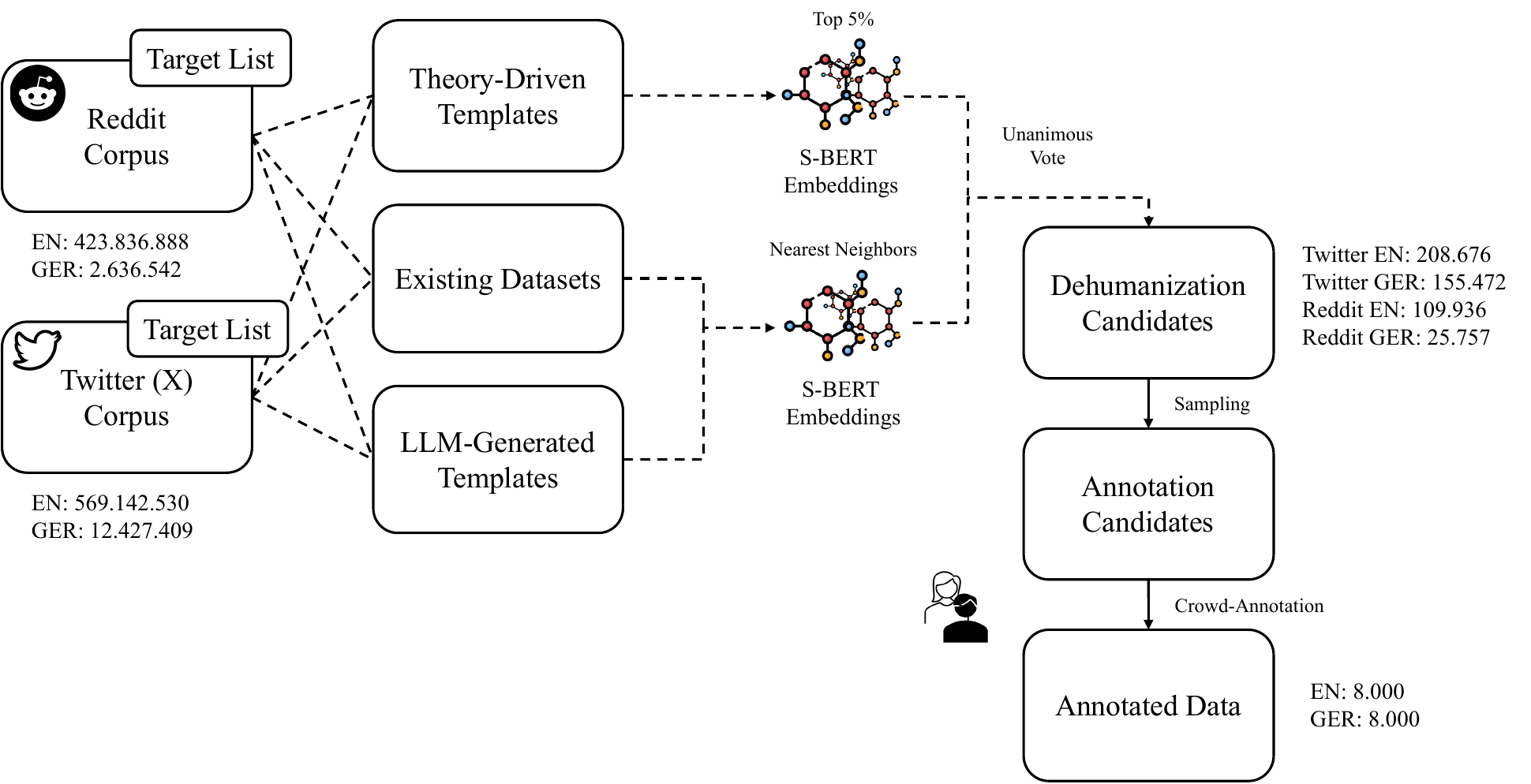}
    \caption{Multi-step pipeline for selecting potential dehumanizing instances for annotation. It includes data filtering based on predefined targets, personal  and demonstrative pronouns, three distinct sampling strategies—theory-informed instance selection, retrieval from existing datasets, and LLM-generated instance matching—and the final annotation process via crowd-workers on Prolific.}
    \label{fig:enter-label*}
\end{figure*}

\subsection{Data Sampling for Annotations}
\label{sec:sampling}
After filtering candidate messages from Twitter and Reddit that potentially refer to targets, we apply three different sampling methods to select instances for human annotations. The goal of our sampling strategies is to increase the prevalence of different forms of dehumanization in our dataset that humans annotate. 


\textbf{Theory-informed Instance Selection:}
We utilize a theory-informed selection method based on survey-scale items designed to measure various dimensions of dehumanization \cite{rasset2023only,kuljian2023warmth,haslam2008attributing}. These survey items, combined with the previously identified frequent targets of dehumanization, are used to create template instances \footnote{Templates are available upon request}. These templates serve as a reference to find semantically similar instances within our real-world datasets from the two platforms. For example, a survey item proposed by \citeauthor{kuljian2023warmth} for measuring implicit animalistic dehumanization is:
\begin{align*}
   \text{[TARGETS] are \textcolor{red}{childish} in their decision making}
\end{align*}

Here, the target is highlighted in blue, and the denial of the human trait---in this case, matureness---is highlighted in red, capturing a subtle form of animalistic dehumanization. Using this approach, we generate and share a comprehensive set of templates that cover all combinations of targets and denial of human traits across various dehumanization dimensions. 
We use three different SBERT models \cite{reimers-2019-sentence-bert} to identify real Twitter or Reddit posts that are semantically similar to predefined templates. For each instance in our dataset, we generate a sentence embedding and compute its cosine similarity with all template embeddings, ranking instances by their similarity scores for each model. We then select the top 0.5\% of instances per model, resulting in 380.000 potential candidates. 



\textbf{Exisiting Dataset Instance Selection}
As a second strategy for instance selection, we utilize existing dehumanization datasets to identify semantically similar examples within our corpus.  This approach helps identify examples that do not closely match our predefined theory-informed templates. Specifically, we collect all dehumanizing instances from publicly available english datasets \cite{kirk-etal-2023-semeval,vidgen-etal-2021-learning,vidgen-etal-2020-detecting,sachdeva-etal-2022-measuring} and identify the three nearest neighbors in our corpus, explicitly excluding exact duplicates. Since existing datasets focus mainly on explicit animalistic dehumanization dimension we expect to find instances that also reflect this dimension. 

\textbf{LLM Instance Selection}
In our third instance selection approach, we utilize \texttt{Llama 3.1} to generate artificial instances by prompting the model to create realistic examples for each dimension of dehumanization \footnote{We share the artificial candidates upon request}. We experiment with various strategies to bypass the model’s guardrails that prevent it from producing harmful content. We use these synthetic examples solely for information retrieval, aiming to identify nearest-neighbor real-world posts within our dataset using SBERT. As our datasets only include data up to 2022, we do not anticipate a significant presence of LLM-generated content.

\paragraph{Selecting Candidates for annotation}

Using our three different instance selection approaches, we gathered 364,148 potential Twitter and 135,693 potential Reddit candidates for crowd annotation. From these, we sampled for each of the two languages  4,000 potentially explicit and 4,000 potentially implicit instances for annotation (weighted sampling according to their similarity scores).

\begin{table*}[ht]
    \small
    \centering
    \renewcommand{\arraystretch}{1.1}
    \setlength{\tabcolsep}{7pt}
    \resizebox{\textwidth}{!}{
    \begin{tabular}
        {l@{\hspace{1\tabcolsep}}r rr rr rr rr rr rr}
        \toprule
        & & \multicolumn{6}{c}{\textbf{\thead{English}}} & \multicolumn{6}{c}{\textbf{\thead{German}}} \\
        \cmidrule(lr){3-8} \cmidrule(lr){9-14}
        \multicolumn{2}{l}{\multirow{3}{*}{\textbf{\thead{Dehumanization}}}} & \multicolumn{2}{c}{\textbf{\thead{Twitter}}} & \multicolumn{2}{c}{\textbf{\thead{Reddit}}} & \multicolumn{2}{c}{\textbf{\thead{Total}}}& \multicolumn{2}{c}{\textbf{\thead{Twitter}}} & \multicolumn{2}{c}{\textbf{\thead{Reddit}}} & \multicolumn{2}{c}{\textbf{\thead{Total}}} \\
        & & \multicolumn{1}{c}{\textbf{\#}} & \multicolumn{1}{c} {\textbf{\%}} & \multicolumn{1}{c}{\textbf{\#}} & \multicolumn{1}{c} {\textbf{\%}} & \multicolumn{1}{c}{\textbf{\#}} & \multicolumn{1}{c} {\textbf{\%}} & \multicolumn{1}{c}{\textbf{\#}} & \multicolumn{1}{c} {\textbf{\%}} & \multicolumn{1}{c}{\textbf{\#}} & \multicolumn{1}{c} {\textbf{\%}} & \multicolumn{1}{c}{\textbf{\#}} & \multicolumn{1}{c} {\textbf{\%}} \\
        \midrule   
        \multicolumn{2}{l}{\texttt{Total}} & \multicolumn{2}{c}{\num{2221}} & \multicolumn{2}{c}{\num{1779}} & \multicolumn{2}{c}{\num{4000}} & \multicolumn{2}{c}{\num{3520}} & \multicolumn{2}{c}{\num{477}} & \multicolumn{2}{c}{\num{4000}} \\
        \cmidrule(lr){3-4} \cmidrule(lr){5-6} \cmidrule(lr){7-8} \cmidrule(lr){9-10} \cmidrule(lr){11-12} \cmidrule(lr){13-14}
        \multicolumn{2}{l}{\texttt{Not Dehumanizing}} & \num{1257} & \num{56.60}\% & \num{1152} & \num{64.76}\% & \num{2409} & \num{60.22}\% & \num{2370} & \num{67.3}\% & \num{351} & \num{73.6}\% & \num{2721} & \num{68.0}\% \\ 
        \multicolumn{2}{l}{\texttt{Explicit}} & \num{964} & \num{43.40}\% & \num{627} & \num{35.24}\% & \num{1591} & \num{39.77}\% & \num{1150} & \num{32.7}\% & \num{126} & \num{26.40}\% & \num{1276} & \num{32.7}\% \\ 
        & \texttt{Animalistic} & \num{725} & \num{32.64}\% & \num{458} & \num{25.74}\% & \num{1183} & \num{29.57}\% & \num{893} & \num{25.37}\% & \num{100} & \num{21.00}\% & \num{993} & \num{24.80}\% \\
        & \texttt{Mechanistic} & \num{312} & \num{14.05}\% & \num{196} & \num{11.02}\% & \num{508} & \num{12.70}\% & \num{209} & \num{6.00}\% & \num{24} & \num{5.00}\% & \num{233} & \num{5.80}\% \\
        \cmidrule(lr){3-8} \cmidrule(lr){9-14}
        \multicolumn{2}{l}{\texttt{Offensive}} & \num{1329} & \num{59.84}\% & \num{873} & \num{49.07}\% & \num{1063} & \num{53.15}\% & \num{2202} & \num{55.05}\% & \num{117} & \num{45.00}\% & \num{1126} & \num{56.30}\% \\ 
        \midrule 
        \multicolumn{2}{l}{\texttt{Total}} & \multicolumn{2}{c}{\num{1558}} & \multicolumn{2}{c}{\num{2442}} & \multicolumn{2}{c}{\num{4000}} & \multicolumn{2}{c}{\num{3458}} & \multicolumn{2}{c}{\num{542}} & \multicolumn{2}{c}{\num{4000}} \\
        \cmidrule(lr){3-4} \cmidrule(lr){5-6} \cmidrule(lr){7-8} \cmidrule(lr){9-10} \cmidrule(lr){11-12} \cmidrule(lr){13-14}
        \multicolumn{2}{l}{\texttt{Not Dehumanizing}} & \num{549} & \num{35.24}\% & \num{1089} & \num{44.59}\% & \num{1638} & \num{40.95}\% & \num{1773} & \num{51.27}\% & \num{364} & \num{67.16}\% & \num{2137} & \num{53.42}\% \\        
        \multicolumn{2}{l}{\texttt{Implicit}} & \num{1009} & \num{64.76}\% & \num{1353} & \num{55.41}\% & \num{2362} & \num{59.05}\% & \num{1685} & \num{48.73}\% & \num{178} & \num{32.84}\% & \num{1863} & \num{46.58}\%\\
        & \texttt{Animalistic} & \num{640} & \num{41.78}\% & \num{731} & \num{29.93}\% & \num{1371} & \num{34.27}\% & \num{1112} & \num{32.16}\% & \num{67} & \num{12.36}\% & \num{1179} & \num{29.47}\% \\ 
        & \texttt{Mechanistic} & \num{369} & \num{23.68}\% & \num{622} & \num{25.47}\% & \num{991} & \num{24.77}\% & \num{573} & \num{16.57}\% & \num{111} & \num{20.48}\% & \num{684} & \num{17.10}\% \\ 
        \cmidrule(lr){3-8} \cmidrule(lr){9-14}
        \multicolumn{2}{l}{\texttt{Offensive}} & \num{886} & \num{56.87}\% & \num{1123} & \num{45.99}\% & \num{2009} &\num{50.22}\% & \num{1794} & \num{51.88}\% & \num{181} & \num{33.39}\% & \num{1975} & \num{49.37}\% \\
        \bottomrule
    \end{tabular}
    }
    \caption{Dataset description. The explicit and implicit dataset consist each of \num{4,000} instances, both categorized through span-annotations and majority voting. 
    Additionally, \texttt{Offensive} annotations were made independently on a document-level}
    \label{tab:dataset-description}
\end{table*}

\subsection{Annotation}
The data extracted through the process described in Section~\ref{sec:sampling} consists of potential dehumanizing text instances. We collected span-level and document-level annotations\footnote{The annotation task interface can be seen in Appendix \ref{app:task}} for a subset of these instances using Prolific, a crowd-working platform, based on the taxonomy shown in Figure~\ref{fig:smart_card_types}. Before launching the main annotation study, we conducted several pilot studies (in our lab as well as on prolific) that tested different task designs, instructions, and user interfaces. These pre-studies helped us assessing  task clarity, annotator agreement, and efficiency. Based on their outcomes, we selected the most effective setup for the main round.

Each instance was annotated by three independent crowd-workers. To ensure annotation quality, we required annotators to be fluent in English (or German), familiar with social media platforms, and hold at least a high school degree or equivalent. For the English task, we required crowdworkers to have an acceptance rate above 98\%. For German, this criterion could not be applied due to the limited availability of native-speaking annotators. Annotators were asked to perform the following three tasks:

\begin{enumerate}
\item	Determine whether the post was perceived as an offensive statement directed at an individual or group.
\item	Highlight any portion of the text that exhibits animalistic or mechanistic dehumanization (see Appendix \ref{app:guidelines} for definitions and examples).
\item	Identify and highlight the target of the dehumanizing expression.
\end{enumerate}

\paragraph{Annotation Results}

Table \ref{tab:dataset-description} presents statistics for the annotated datasets.  Additionally, we report the Inter-Annotator Agreement (IAA), measured using Krippendorff’s $\alpha$. For explicit dehumanization, the English dataset exhibits an IAA of 0.393 ($\alpha$ = 0.475 for German data). Looking at the theoretical subdimensions of dehumanization reveals that the IAA for English explicit animalistic dehumanization is slightly higher (0.431). Specifically, the animal comparison category shows high agreement, with an IAA of 0.577. In contrast, subhuman, one of the most complex subcategories, has the lowest agreement at 0.275. For English explicit mechanistic dehumanization, the agreement is slightly lower, with $\alpha$ at 0.381. For implicit animalistic dehumanization, $\alpha$ is 0.290 for English and 0.517 for German, whereas it is 0.292 (English) and 0.289 for implicit mechanistic dehumanization. These relatively low scores have been observed in previous studies on dehumanization \cite{sachdeva-etal-2022-measuring} and can be attributed to several factors. First, the perception of dehumanization is inherently subjective. Individual annotators may vary in their sensitivity to particular expressions, depending on their background, values, or lived experiences. Second, without sufficient context, it is often difficult to determine whether a statement is intended as dehumanizing. For example, the statement \textit{``Not, but see because they’re actually maggots over there''} could either imply a comparison between people and maggots or simply describe the presence of maggots in a certain location, making its interpretation ambiguous. We emphasize that this annotation task is inherently challenging, particularly because all instances were selected as already being potential candidates for dehumanization. Moreover, \citet{wong-etal-2021-cross} argue that in tasks involving subjective judgment, like identifying dehumanizing language, a Cohen's kappa threshold of 0.6 is too strict, since annotators often vary in cultural background and training. 
For a stratified random sample used as test data in our experiments, three experts (authors of the paper) annotated the instances with the same instructions as the crowdworkers, achieving a level of agreement comparable to that of the crowdworkers. However, unlike the crowdworkers, the experts resolved their disagreements through several rounds of discussion. 
The IAA of the crowdworker can be found in the the Appendix \ref{sec:appendix_iaa}. 


\paragraph{Candidate Selection Results }

\begin{figure}[ht]
    \centering
    \begin{subfigure}[t]{0.4\textwidth}
        \centering
        \includegraphics[width=\linewidth]{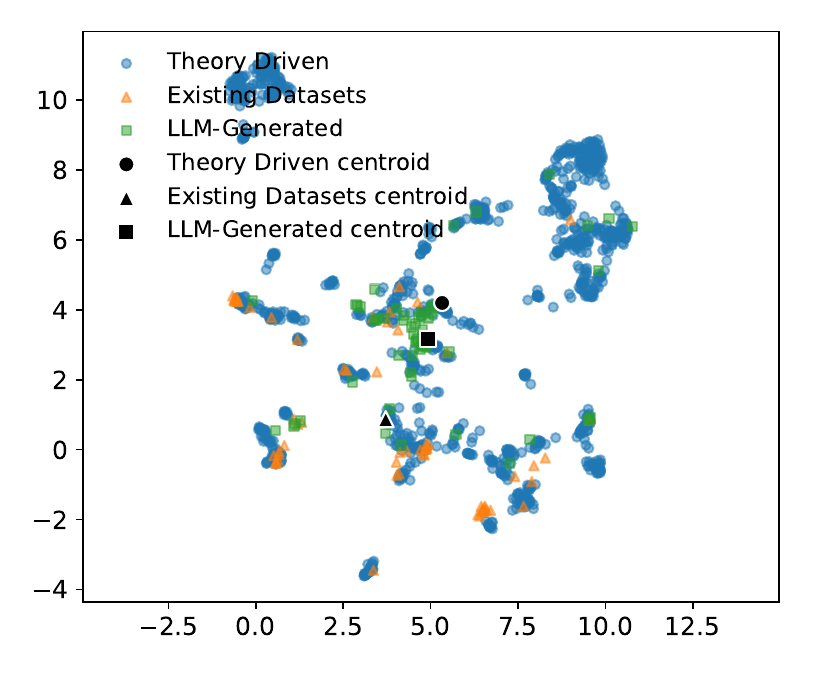}
        \caption{Explicit candidates}
        \label{fig:sub1}
    \end{subfigure}
    \hfill
    \begin{subfigure}[t]{0.4\textwidth}
        \centering
        \includegraphics[width=\linewidth]{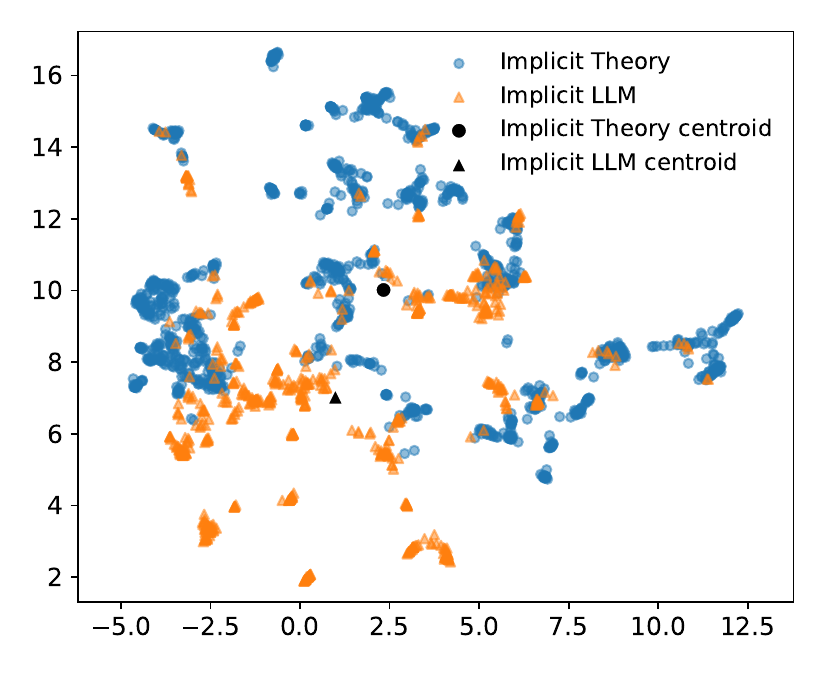}
        \caption{Implicit candidates}
        \label{fig:sub2}
    \end{subfigure}
    \caption{Semantic space of (a) explicit and (b) implicit candidates that are positively labeled by the crowd as dehumanization.}
    \label{fig:diversity}
\end{figure}

Table \ref{tab:dehumanization-distribution} reveals for the different candidate selection strategies the fraction of candidates that were labeled as dehumanization by the majority of our annotators.  Note that we have not used existing datasets to select implicit candidates as the ones that we use only focus on the explicit dehumanization dimension. 

Table \ref{tab:dehumanization-distribution} shows that the different sampling methods differ in their ability to find dehumanizing examples in the four relevant categories. Theory-informed candidate selection provides the highest fraction of positive instances for all categories, except for implicit mechanistic dehumanization. Here, LLM-based templates are more efficient in finding relevant instances in the real-world data.

Additionally, we examined the diversity of instances positively annotated as dehumanization, originating from different candidate selection approaches. We calculated the pairwise cosine distances between their sentence embeddings. For the explicit dimension, each candidate selection strategy produced a relatively diverse set of candidates, with an average intra-strategy cosine distance of 0.79 and 0.80 for implicit instances. A visual representation of the sentence embeddings, projected into a two-dimensional space using UMAP, supports the finding that all sampling strategies find relatively diverse set of instances (cf. Figure \ref{fig:diversity}).



\begin{table}[htbp]
\centering
\small
\begin{tabular}{lrrr}
\toprule
\textbf{Category} & \textbf{Theory} & \textbf{LLM} & \textbf{Existing} \\
\midrule
Exp Animal & 54\% & 21\% & 16\% \\
Exp Mecha & 22\% & 8\% & 5\% \\
Imp Animal& 75\% & 54\% & - \\
Imp Mechan & 40\% & 65\% & - \\
\bottomrule
\end{tabular}
\caption{Fraction of dehumanizing instances retrieved with a sampling method (column) for a specific dimension of dehumanization (row). For example, one can see that 54\% of  all explicit animalistic dehumanization candidates that were retrieved with theory-informed candidate selection are indeed dehumanizing according to the majority vote of annotators.}
\label{tab:dehumanization-distribution}
\end{table}


\section{Experimental Setup}

We develop eight separate models, corresponding to the binary explicit/implicit and mechanistic/animalistic dehumanization dimensions in English and German, and evaluate their performance on fixed hold-out splits. Therefore, we aggregate the span-level annotations into document-level annotations that correspond to the four dimensions of dehumanizations outlined in Haslam's Dual Dehumanization model \cite{haslam-review}. For example, within the explicit dehumanization
task annotators had to decide if a message compares a target group to an animal, a disease,
a subhuman creature, or an inanimate object. If nothing applied the message is considered
as “not explicitly dehumanizing”. If an animal, disease or subhuman creature comparison is
highlighted the message is considered “explicit animalistic dehumanization”. If a comparison with an inanimate object is found, the instance is an example of “explicit mechanistic
dehumanization”. In theory one instance can contain multiple forms of dehumanization (e.g.,
animalistic and mechanistic dehumanization), but this did not happen in practice. Ties (e.g.,
animal comparison vs. inanimate comparison vs. no comparison) were rare and were resolved
manually.
We rely on methods that are frequently used for hate speech detection such as fine-tuned BERT (\texttt{bert-base-uncased} for English and \texttt{bert-base-multilingual-uncased} for German) \cite{DBLP:journals/corr/abs-1810-04805} and RoBERTa (\texttt{xml-roberta-large}) \cite{DBLP:journals/corr/abs-1911-02116}  models, alongside established hate speech detection architectures HateBERT \cite{caselli-etal-2021-hatebert} and MetaHateBERT \cite{Piot_Martin-Rodilla_Parapar_2024}\footnote{As dehumanization is a subconstruct of hate speech we expect that these models are able to capture some of its signals}. Among the established hate speech detection models, both were fine-tuned on our dataset for 3 epochs with a learning rate of 5e-5, batch size of 32, weight decay of 0.18, and Adam 8bit optimizer. 

In addition to our experiments with BERT-based model architectures, we also tested generative large language models based on a decoder-only architecture. We examined Llama 3.1 8B (\texttt{Llama-3.1-8B-Instruct}) and Llama 3.1 70B (\texttt{Llama-3.1-70B-Instruct}) \cite{dubey2024llama3herdmodels} in zero-shot and few-shot (20 shots) settings with the same instructions given to the human annotators (see Appendix \ref{sec:appendix}). Furthermore, we fine-tuned the LLMs via Low-Rank Adaptation \cite{lora} for 4 epochs using a learning rate of 2e-4, batch size of 4, weight decay of 0.01 and AdamW torch optimizer. For the german explicit dimension we undersampled the data because of the low number of positive instances. Evaluation was performed using greedy decoding (temperature = 0, no sampling), such that the model always produced the most likely continuation. Finally, we use a black-box model, the Perspective API (toxicity score),  that we do not fine-tune or train on our data \cite{10.1145/3534678.3539147}. We also test the performance of MetaHateBERT and HateBERT without fine-tuning them.



\begin{figure*}[htb]
  \centering
  \includegraphics[width=\textwidth]{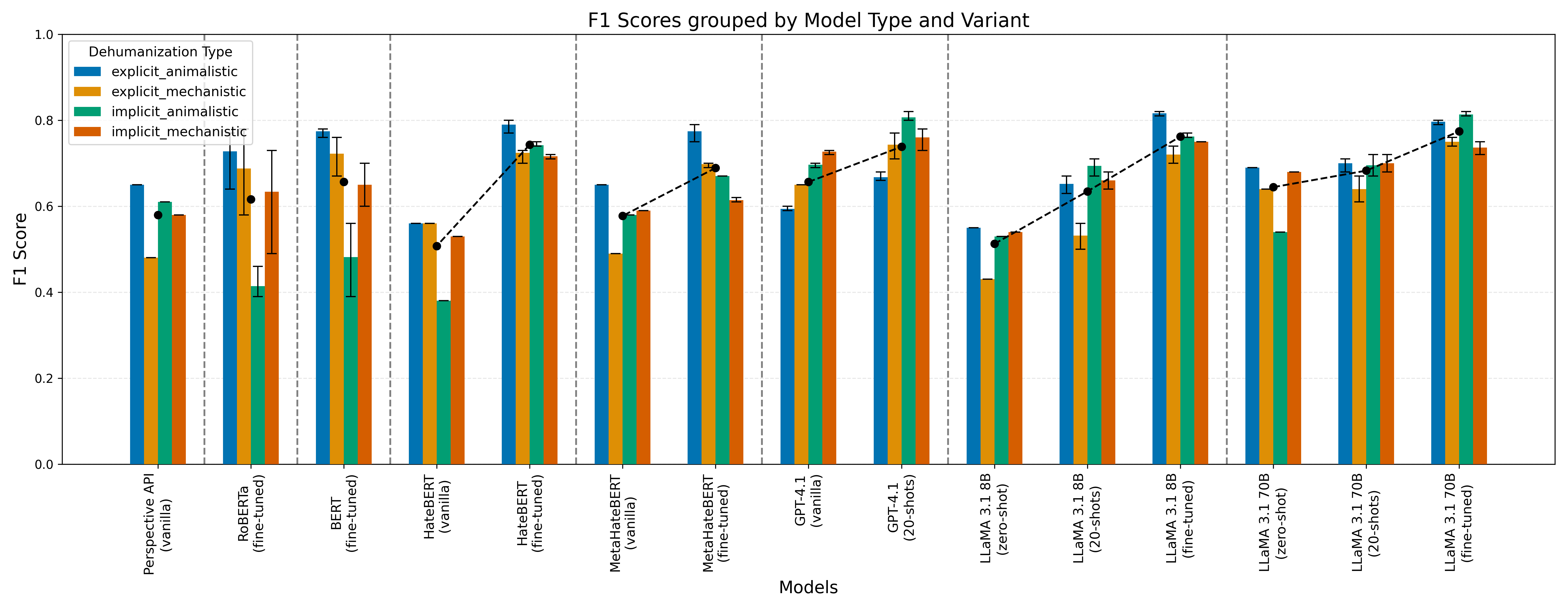}
  \caption{Test F1-Scores for the four different dimensions of dehumanization (see color bars) on English Reddit and Twitter data for different model architectures. }
  \label{fig:f1_scores_grouped}
\end{figure*}

\section{Results}

The comparison of different models and model architectures, as shown in Figure~\ref{fig:f1_scores_grouped}, reveals that all evaluated models benefit from training on our dataset. These improvements are consistent across both few-shot and finetuning settings. Notably, we observe a stepwise increase in performance: from zero-shot to few-shot to fine-tuned models, with each level of supervision yielding better results. Corresponding results for the German dataset are provided in Figure~\ref{fig:_ger_f1_scores_grouped} in the Appendix and follow a similar trend. In the following, we discuss general findings that apply to both languages.

\paragraph{Out-of-the-Box Performance}
Models that have not been exposed to any data from our dataset provide a useful baseline for evaluation. These include both models pretrained on related hate speech or abusive language tasks—such as \texttt{Perspective API}, \texttt{HateBERT} (vanilla), and \texttt{MetaHateBERT} (vanilla)—and general-purpose language models like \texttt{Llama 3.1} (zero-shot) and \texttt{GPT-4.1}.

Baseline toxicity classifiers, such as the Perspective API, HateBERT, and MetaHateBERT, perform reasonably well on instances of animalistic dehumanization with strong negative valence — for example: \textit{``[USER] Indians are like virus. They spread so quickly and are most unwanted everywhere...''}. However, these models tend to fail when the dehumanization is more implicit and lacks explicitly negative wording. For example \textit{``My dogs are Republicans too, when one barks for no reason they all bark.''}, where the comparison subtly implies that Republicans exhibit mindless herd behavior. Such indirect and figurative expressions often go undetected by these standard toxicity models whereas they can be detected with fine-tuned LLMs.


In the zero-shot setting, \texttt{GPT-4.1} outperforms the open-source \texttt{Llama 3.1} variants but still shows limited overall performance. Its only notable success appears in the implicit mechanistic category, where it achieves a relatively high F1 score of 0.72.

\paragraph{Few-Shot and Fine-Tuned Models}

Models trained on our dataset show substantial performance gains across all dehumanization categories. Even in the few-shot setting, where only a small number of examples are provided in the prompt, performance improved by an average of 12\% compared to the corresponding zero-shot scenarios. The best average results across all four dimensions were achieved by the fine-tuned \texttt{Llama 3.1-Instruct} models (Llama 70B fintuned $F1=0.774$ and Llama 8B fine-tuned $F1=0.76$, followed by fine-tuned HateBERT $F1=0.74$ and GPT-4.1 few shot $F1=0.73$).
Finetuning on our data proved beneficial across all dimensions, with the largest gains observed for the implicit and explicit animalistic categories. The former benefits from learning to recognize indirect animal comparisons, while the latter improves due to the model's exposure to more diverse examples of denials of unique human traits.

While the larger 70B Llama model does outperform the smaller 8B variant, the performance difference is relatively modest in the fine-tuning setting and for some dimensions (e.g. explicit animalistic) the 8B fine-tuned model even outperforms its larger counterpart. This result is particularly relevant for practical applications, as the 8B model is significantly more cost-efficient, environmentally friendly, and easier to deploy. It is therefore a viable option for scenarios with limited computational resources~\cite{Piot2025}. However, in the zero-shot setting, the larger model variants exhibit a clear advantage, suggesting that they possess a stronger inherent understanding of the underlying constructs and therefore achieve higher performance without task-specific supervision.

\paragraph{Error Analysis}

We conducted a qualitative analysis to better understand the types of errors made by our models. Specifically, we used our complete test data (800 instances for each dimension) with the agreed-upon expert labels and examined true positives (TP), true negatives (TN), false positives (FP), and false negatives (FN) that were consistently shared across the majority of our fine-tuned models (Llama, HateBERT, MetaHateBERT).

For the correctly classified instances, we observe that our fine-tuned models appear to have learned not to rely solely on specific trigger words (like the mentioning of an animal as an indicator for explicit animalistic dehumanization) to assign positive class labels, thereby avoiding spurious correlations. This suggests that the models are not merely memorizing surface-level lexical cues, but instead developing a more contextualized understanding of dehumanizing language. For instance, sentences like
\textit{``There are no male or female Earthworms. They’re just \ Earthworms.''}
are correctly classified as non-dehumanizing, as the reference clearly pertains to actual animals rather than metaphorical or figurative language aimed at humans. This indicates that the models can distinguish between literal and metaphorical usage of animal terms—an important capability for avoiding false positives in real-world applications.

In addition to analyzing correctly classified instances, we conducted a systematic investigation of the model’s error modes by qualitatively examining misclassified examples and deriving distinct categories of errors. For the explicit dimensions, we observe that false positives mainly originate from ambiguous instances that can be interpreted in multiple ways. Statements like \textit{”[USER] yeah so I’m calling it the Chinese Virus again”} may be perceived as a comparison of Chinese people to a virus, and thus as a form of explicit dehumanization. However, they can also be interpreted as politically charged rhetoric referring to the origin of the virus in China, rather than targeting individuals or groups directly. On the other hand, we observe a category of false positives where the model predicts animalistic dehumanization, likely falling into the subhuman subtype, while annotators labeled the instance as unproblematic. For example, the statement \textit{“That’s how I really feel. Indians are a filthy people/culture. Do some research, you’ll see.”} contains derogatory generalizations that may lead the model to infer dehumanization. However, annotators have annotated this not as animalistic dehumanization, possibly viewing it as a general expression of prejudice or offensive speech rather than an explicit comparison to animals or subhuman traits. Explicit mechanistic models suffered mostly from false negative classifications specifically in context of sexualized objectification of women as well as instances that described targets as scum or filth. Although these terms are refering to inanimate objects, they were more often annotated by the crowd as belonging to the subhuman rather than the inanimate category. 
For the implicit dimensions we observe that the models struggled (in contrast to the explicit dimensions) with counter-speech e.g. \textit{``All poor people are on welfare and leech off of the hard working Americans.   Yeah, right....''} or \textit{``That would make them brainless sheep, and I don’t think they’re that way at all.''}. Such instances often repeat dehumanizing stereotypes in order to reject them, which requires models to distinguish between reinforcement/endorsement and negation of harmful generalizations. Also for the implicit dimensions, we observe that a substantial portion of errors stems from subjective or ambiguous cases that are inherently difficult to classify, even for human annotators. For example, the sentence
\textit{``They look the same as Turks and blend in really well among Turks.''}
could be interpreted as implying that a certain group is \textit{fungible} or \textit{interchangeable}, which aligns with mechanistic dehumanization. Without broader context—such as intent, tone, or surrounding discourse—interpretations remain speculative, increasing the risk of misclassification. 

\begin{figure}
    \centering
    \includegraphics[width=0.99\linewidth]{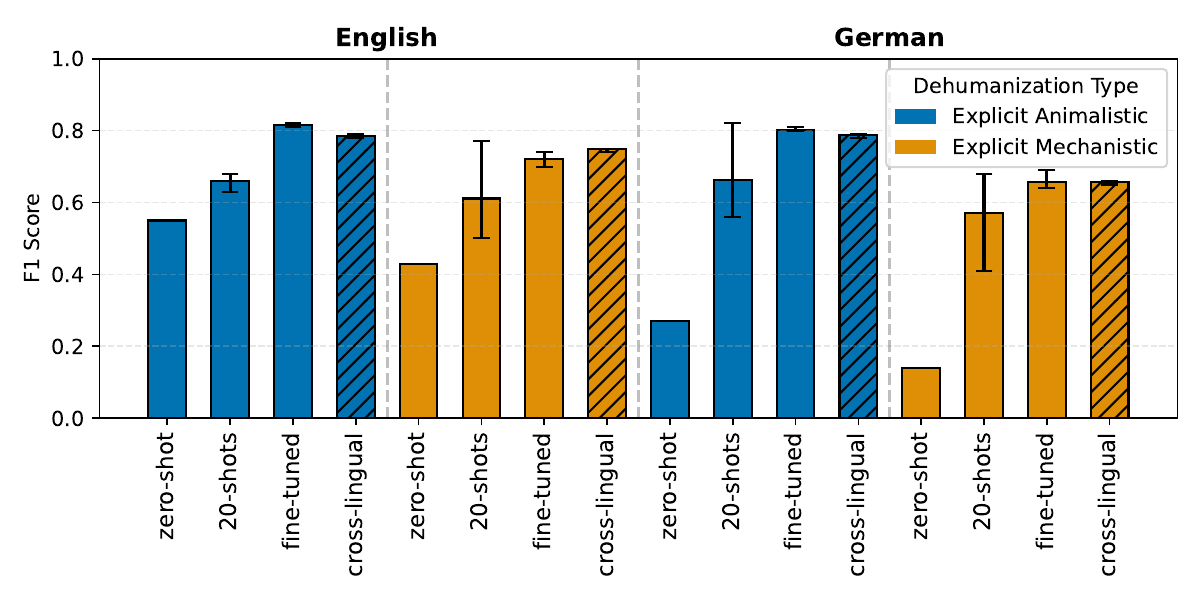}
    \caption{Cross-lingual analysis of explicit dehumanization models. The labels on top indicate the evaluation language. Cross-lingual models (hatched bars)  were fine-tuned on the opposite language.}
    \label{fig:cross-lingual}
\end{figure}

\paragraph{Crosslingual Analysis}
In addition to our monolingual analysis, we evaluated the performance of Llama 3.1 different in cross-lingual settings. Specifically, we tested whether models fine-tuned on data from one language can generalize to another. Figure~\ref{fig:cross-lingual} presents results for the explicit animalistic and mechanistic dehumanization dimensions. The findings demonstrate that our annotated data enables robust model performance even in cross-lingual applications, suggesting generalizability across linguistic and cultural boundaries. Notably, models fine-tuned on a different language (represented by hatched bars) performed comparably to models fine-tuned on the evaluation language and outperformed few-shot baselines. Interestingly, the explicit mechanistic model fine-tuned on English data outperforms the model fine-tuned on German data when evaluated on the German test set. We attribute this to the smaller number of positive training instances available in the German dataset compared to the English dataset (see Table \ref{tab:dataset-description}).


\section{Discussion}
The review of existing approaches to measuring dehumanizing content revealed several limitations, most notably the scarcity of datasets, their narrow focus on explicit, often animalistic comparisons, and their focus on English language. To address this gap, we collected a large, bilingual dataset from $\mathbb{X}$ (former Twitter) and Reddit that covers the different theoretical dimensions of dehumanization  and stretches over more than one decade. 


Our dataset includes both dehumanizing and non-dehumanizing examples, often topically or linguistically similar due to a similarity-based retrieval strategy. Consequently, even negatively labeled examples are plausible dehumanization candidates, making the dataset challenging and valuable for both training and rigorous evaluation.
Our baseline experiments show that current general-purpose LLMs, as well as classifiers pretrained on abusive language and hate speech tasks, struggle to adequately identify the different dimensions of dehumanization and benefit from utilizing our dataset in few-shot or finetuning scenarios.
An additional advantage of our dataset is the inclusion of span-level annotations, allowing for a more precise identification of dehumanizing language within text that can be used in the future. We also publish the full set of dehumanization candidates (over 490,000), along with the corresponding predictions and logits made by our best-performing models (see Appendix \ref{app:candidate_data_predictions}). This dataset may serve as a valuable resource for selecting new instances for annotation in the future.

We compare the performance of 14 different models on the binary dehumanization detection for each sub-dimension of the construct. Our results show that most methods (except GPT-4.1) show their best performance on explicit animalistic dehumanization in the English data. In the German data we see that  explicit and implicit animalistic dehumanization are the best performing dimensions for all models including GPT-4. In general, vanilla and zero-shot models show much lower performance than fine-tuned or few-shot models. The only exception here is GPT-4 which also show good performance in the zero-shot setting. We chose not to fine-tune GPT-4  since as a proprietary model, it does not offer transparency into the fine-tuning process. This limits reproducibility and interpretability. In addition, we aim to provide the model with as little information as possible to avoid data leakage. However, we find that all models trained on our dataset show substantial performance gains across all dehumanization categories. Even in the few-shot setting, where only a small number of examples are provided in the prompt, performance improved by an average of 12\% compared to the corresponding zero-shot scenarios. Our preliminary cross-lingual analysis also demonstrates that our annotated data enables robust model performance even in cross-lingual applications.



\section{Limitations}
We aimed to create a comprehensive dataset by incorporating a wide range of dimensions and frequently targeted groups that goes beyond specific target groups that are addressed in current research. However, we cannot ensure that all possible targets of dehumanization are included and our dataset is limited to the list of target identifiers that we used. To address this limitation we complemented the target list with personal pronouns like “they” and  demonstrative pronouns like “those” and “these” that imply group references.

Our dataset is also limited to two platforms (Twitter and Reddit) and two languages (German and English) and is not a random or representative sample of the content that was published at these platforms between 2013 and 2023. However our dataset provides a diverse set of examples of dehumanizing messages that were published over the course of ten years.


Modeling the explicit mechanistic dimension, especially in German, was challenging due to a lack of training instances, likely reflecting the lower prevalence of mechanistic dehumanization compared to animalistic forms. Future efforts should focus on gathering more examples in this category.

Furthermore, dehumanization often occurs not only through text, but also through images or memes. Since our dataset includes only textual content, we capture text-based signals only.

\bibliography{custom}

\bibliographystyle{acl_natbib}

\appendix
\onecolumn

\section{Target Selection}
\label{sec:appendix}

\begin{table}[H]
    \footnotesize
    \centering
    \begin{tabular}{p{1.7cm} p{13.4cm}}
        \toprule
        \textbf{\small Target} & \textbf{\small Terms} \\
        \midrule
        Foreigners / Immigrants & immigrant, immigrants, refugee, refugees, alien, aliens, illegal, illegals, asylum, escapee, escapees, foreigner, foreigners, exile, displaced person, displaced persons, migrant, migrants, undesirables \\
        \midrule
        Religion & jew, jews, hebrew, hebrews, kike, kikes, yid, yids, mockie, mockies, mocky, mockys \\
        \midrule
        Race & black, blacks, bootlip, bootlips, dindu, dindus, porch monkey, porch monkeys, ghetto monkey, ghetto monkeys, golliwog, golliwogs, hoodrat, hoodrats, jigga, jigger, jigro, kneegrow, kneegrows, moulie, moulies, negro, negros, nigger, niggers, afro-american,afro-americajns, afromerican,aformericans, native-american, native-americans \\
        \midrule
        Gender / Sex / Identity & woman, women, girl, girls, bitch, bitches, cunt, cunts, dyke, dykes, pussy, pussies, hoe, hoes, bint, bints, female, females, shemale, shemales, trans, twat, twats, asexual, asexuals, faggot, faggots, homosexual, homosexuals \\
        \midrule
        Ethnicity / Nationality & japanese, german,  germans, american, americans, asian, asians, turk, turks, arab, arabs, chinese, mexican, mexicans, african, africans, indian, indians, italian, italians, spanish, russian, russians, scot, scots, french, syrian, syrians, dutch \\
        \midrule
        Political & democrat, republican, republicans, democrats, dems, reps \\
        \midrule
        Disability & cripple, cripples, gimp, gimps, libtard, libtards, mongs, mongos, retard, spaz, tard, tarded, tards, mongo, mongos \\
        
        \midrule
        Old people & elderly, elderlies, olderwoman, olderwomans, olderman, oldermen, oldster, oldsters, oldppl, old person, old persons  \\
        
        \midrule
        Medical and criminal & patient, patients, criminal, criminals, inmate, inmates, prisoner, prisoners \\
        
        \bottomrule
    \end{tabular}
    \caption{List of target categories that are frequently reported to be dehumanized. }
    \label{tab:terms}
\end{table}

\begin{table}[H]
    \footnotesize
    \centering
    \begin{tabular}{p{1.7cm} p{13.4cm}}
        \toprule
        \textbf{\small Target} & \textbf{\small Terms} \\
        \midrule
        Foreigners / Immigrants & Einwanderer, Immigrant, Immigranten, Flüchtling, Flüchtlinge, Ausländer, Illegale, Illegaler, Vertriebener, Vertriebene, Unerwünschte, Migrant, Migranten \\
        \midrule
        Religion & Jude, Juden, Hebräer, Jews, Jew, Israeli, Israelis, Palästinenser \\
        \midrule
        Race/Ethnicity & Schwarzer, Schwarze, Neger, Nigger, Niggers, Afroamerikaner, Ureinwohner Amerikas, Chinese, Asiate, Asiaten, Araber, Türke, Türken, Mexikaner, Afrikaner, Inder, Italiener, Spanier, Russe, Russen, Schotte, Schotten, Franzose, Franzosen, Syrer, Niederländer, Ukrainer \\
        \midrule
        Gender / Sex / Identity & Frau, Frauen, Mädchen, Schlampe, Schlampen, Fotze, Fotzen, Lesbe, Lesben, Muschi, Muschis, Hure, Huren, Tussi, Tussis, Weiblich, Weibliche, Transe, Transen, Trans, Transsexuell, Transsexuelle, LGBTQ, Asexuell, Asexuelle, Schwuchtel, Schwuchteln, Homosexuell, Homosexuelle, Schwule, Schwule Menschen \\
        \midrule
        Political & Republikaner, Konservative, Demokrat, Demokraten, Grüne, Ökos, Grünlinge, Sozialdemokraten, SPDler, SPD, CDUler, Christdemokraten, CSUler, CSU, AfDler, Blaue, Rechtskonservative, Liberale, FDPler, Gelbe, Linke, Politiker \\
        \midrule
        Disability & Krüppel, Behinderter, Behinderte, Libtard, Libtards, Mongo, Mongos, Spasti, Spasten, Tard, Tards, Behindert \\
        \midrule
        Old people & Ältere/r, Ältere, Ältere Frau, Ältere Frauen, Älterer Mann, Ältere Männer, Senior, Senioren, Alte Leute, Alte Person, Alte Personen \\
        \midrule
        Medical and criminal & Patient, Patienten, Krimineller, Kriminelle, Diebe, Straftäter, Häftling, Häftlinge, Gefangener, Gefangene \\
        \midrule
        Social class / Occupation & Arbeiter, 9-bis-5-Arbeiter, 9-to-5, Tourist, Reisende, Geschäftsmann, Geschäftsmänner \\
        \bottomrule
    \end{tabular}
    \caption{List of frequently dehumanized target groups in German.}
    \label{tab:terms_german}
\end{table}

\section{Annotation Task and Guidelines}
\label{app:guidelines}

In the following we present the annotation task as well as the guidelines that were provided to the crowd-workers for the different dehumanization annotation tasks.

\subsection{Task}
\label{app:task}

\begin{figure}[h!]
    \centering
    \includegraphics[width=0.9\linewidth]{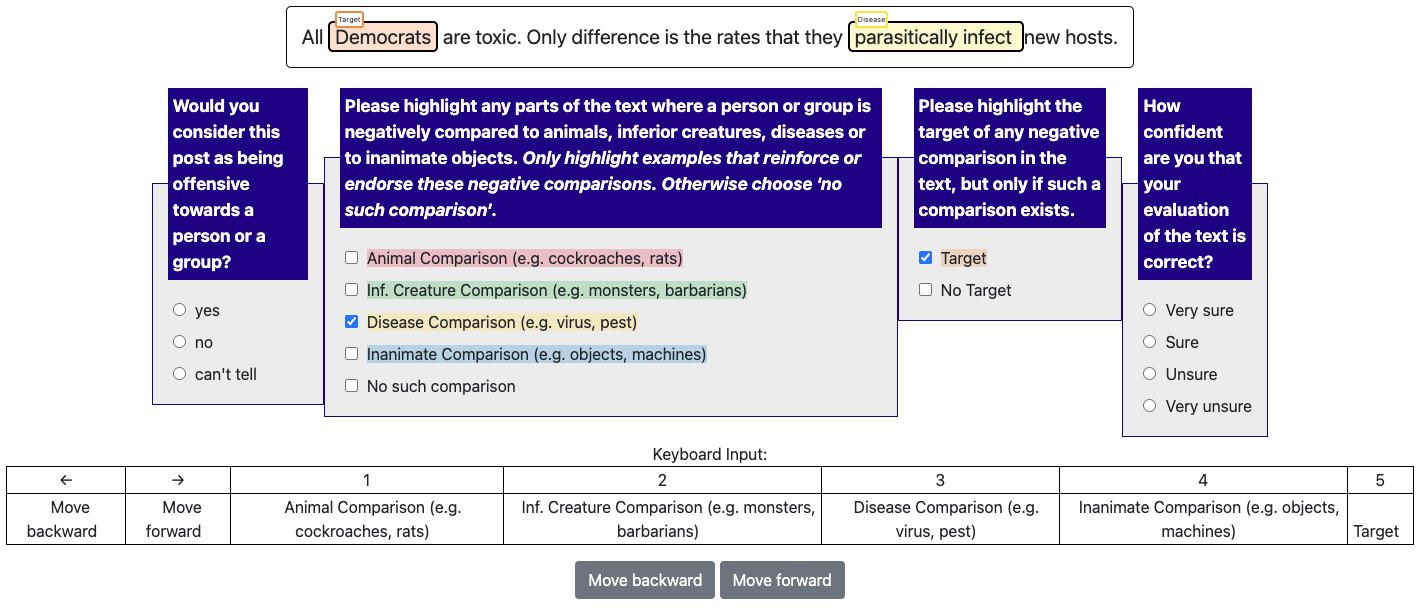}
    \caption{An example of the annotation task that was presented to the crowd-worker (here explicit dehumanization).}
    \label{fig:enter-label}
\end{figure}

\subsection{Explicit Dehumanization}

\paragraph{Explicit Dehumanization Task}
Please read the entire task description and examples carefully!
Your task involves multiple steps.
\begin{itemize}
    \item First, we would like to know whether you perceive a text as generally \textbf{offensive} toward a person or a group of people.
    \item Second, we ask you to find and highlight when the text \textbf{negatively compares} a person or group to \textbf{animals} (e.g., cockroaches, dogs), \textbf{inferior creatures} (e.g., monsters, savages), \textbf{diseases} (e.g., viruses, pests), or \textbf{inanimate objects} (e.g., machines, tools, or robots). We do not distinguish between direct or relative comparisons, so if one group is compared in relation to another group, mark it. Focus only on the text, and ignore any hyperlinks.
\end{itemize}

Only highlight examples that \textbf{reinforce} or \textbf{endorse} these \textbf{negative comparisons}. Do not highlight statements that argue against them. Besides, not all comparisons are harmful—for example, saying someone is ``strong like a lion'' is not dangerous, so do not mark those.

Please highlight both the \textbf{negative comparison} (if present) and the \textbf{target}, including any pronouns referring to the target. Examples are provided below.

\paragraph{Direct Comparison}
\begin{itemize}
    \item That's the thing with the \textbf{Jews}. They're just \textbf{like monkeys} with their lack of knowledge.\\
    \textit{Explanation: This statement directly associates Jewish people with monkeys.}
    \item \textbf{Women} = kitchen, kitchen = dishes, dishes = objects, \textbf{objects = women}.\\
    \textit{Explanation: This statement reduces women to being associated solely with the kitchen and household chores, ultimately equating them with objects.}
    \item \textbf{Republicans} are hateful \textbf{ugly creatures}!\\
    \textit{Explanation: The author of the post directly compares members of the Republican party to creatures.}
    \item Those \textbf{curry boys} are \textbf{filthy little pigs}, so gross.\\
    \textit{Explanation: The statement directly compares a group of people to pigs.}
\end{itemize}

\paragraph{Indirect Comparison through Metaphorical Language and Analogies}
\begin{itemize}
    \item \textbf{A wild flock} of \textbf{Jews} quickly \textbf{flies away}.\\
    \textit{Explanation: Jewish people are implicitly compared to birds through the metaphor of a flock, typically used to describe a group of birds.}
    \item \textbf{She} is my \textbf{possession}. I can use \textbf{her} as I like.\\
    \textit{Explanation: This statement implies that women are considered belongings or property. The metaphorical language used here indirectly associates women with objects that can be owned and controlled, thereby dehumanizing them and reducing their value to that of an item rather than a person.}
\end{itemize}

\paragraph{No (Endorsed) Negative Comparison}
\begin{itemize}
    \item He is strong like a lion.\\
    \textit{Explanation: The statement does not demean the person being compared. Instead, it uses a positive comparison to highlight strength, a trait commonly associated with lions. While it likens someone to an animal, it does so in a way that is complimentary rather than degrading, and therefore does not lower social barriers.}
    \item Who the fuck do you think you are calling migrants cockroaches? Btw, I'm one of those migrants.\\
    \textit{Explanation: The author of the post is not expressing the opinion that migrants are cockroaches. Instead, they are challenging and rejecting this comparison, making it clear that they do not agree with or accept this insult.}
\end{itemize}

\paragraph{No Comparison}
\begin{itemize}
    \item Women. Aren't. Animals. \\
    \textit{Explanation: This statement serves as counterspeech, explicitly rejecting the idea that women should be compared to animals.}
    \item Women love cats.\\
    \textit{Explanation: This is a general statement expressing a common stereotype that some women like cats.}
\end{itemize}

\subsection{Implicit Animalistic Dehumanization}


\paragraph{Implicit Animalistic Task}
Please read the entire task description and examples carefully!
Your task involves multiple steps.

\begin{itemize}
    \item 
    First, we would like to know whether you perceive a text as generally \textbf{offensive} toward a person or a group of people.
    \item Second, we ask you to find and highlight any part of the text that directly or indirectly \textbf{portrays} an \textbf{individual} or a \textbf{group of humans} as being \textbf{incivil/uncultured} (rude, coarse, disrespectful, or lacking in courtesy and politeness), \textbf{irrational} (actions, thoughts, or beliefs that contradict reason or logic), \textbf{unintelligent} (lack of mental ability, understanding and difficulty grasping complex ideas), \textbf{immature/child-like} (lack of emotional development, responsibility, or maturity), or \textbf{lacking morals} (dishonest, unethical, or unscrupulous, prone to violate the moral code). Focus only on the text, and ignore any hyperlinks.
\end{itemize}

Please highlight both the \textbf{comparison} (if present) and the \textbf{target}, including any pronouns referring to the target. Examples are provided below.

\paragraph{Highlights}
\begin{itemize}
    \item Some \textbf{women} these days \textbf{have no morals} whatsoever.\\
    \textit{Explanation: Women are explicitly depicted as \textbf{lacking morals}.}
    \item Most \textbf{Asians} have \textbf{no manners}; it's like they are \textbf{allergic to saying thank you}.\\
    \textit{Explanation: The author claims that Asian people are \textbf{uncultured}.}
    \item \textbf{Women} are actually capable of logical thinking, perhaps \textbf{not to the same degree as men}.\\
    \textit{Explanation: Women are described as being \textbf{irrational}, compared to men.}
    \item \textbf{They} are thinking outside the box, but unfortunately, \textbf{they're not thinking at all}.\\
    \textit{Explanation: The language used suggests that individuals are being characterized as \textbf{unintelligent}.}
    \item Watching \textbf{Republicans} in Congress feels like \textbf{babysitting a room full of toddlers}. \textbf{Tantrums, refusing to share, and making a mess}.\\
    \textit{Explanation: Republicans are portrayed as \textbf{immature} and behaving like toddlers.}
\end{itemize}

\paragraph{No Highlights}
\begin{itemize}
    \item When people say Arabs are rude, I just get a little tired of them.\\
    \textit{Explanation: It is not the opinion of the author that people from a certain ethnic group are rude.}
    \item Irrationality is a curse.\\
    \textit{Explanation: No target group or individual is accused of being irrational.}
    \item I hate these fucking women bitches!\\
    \textit{Explanation: While this sentence is derogatory towards women, the target group is not described as incivil/uncultured, coarse, irrational, unintelligent, immature/child-like, or lacking morals.}
\end{itemize}

\subsection{Implicit Mechanistic Dehumanization}


\paragraph{Implicit Mechanistic Task}
Please read the entire task description and examples carefully!
Your task involves multiple steps.

\begin{itemize}
    \item First, we would like to know whether you perceive a text as generally \textbf{offensive} toward a person or a group of people.
    \item Second, we ask you to find and highlight any part of the text that directly or indirectly \textbf{portrays} an \textbf{individual} or a \textbf{group of humans} as being \textbf{cold} (emotionally distant, unfeeling, or lacking warmth and compassion), \textbf{passive/lacking agency} (accepting or allowing things to happen without active response or resistance, often showing a lack of initiative or assertiveness), \textbf{superficial/lacking depth} (concerned only with surface appearances or shallow aspects, lacking depth, substance, or thoroughness), \textbf{fungible/interchangeable} (someone who is interchangeable or replaceable, typically without any loss of value), or \textbf{rigid/inflexible} (someone who is inflexible, unwilling to change, or firmly fixed in position, behavior, or beliefs, often lacking adaptability). Focus only on the text, and ignore any hyperlinks.
\end{itemize}

Please highlight both the \textbf{comparison} (if present) and the \textbf{target}, including any pronouns referring to the target. Examples are provided below.

\paragraph{Highlights}
\begin{itemize}
    \item Trying to get compassion from \textbf{Republicans} is like \textbf{asking a thermostat for a hug}.\\
    \textit{Explanation: Republicans are described as \textbf{emotionally distant}.}
    \item Some \textbf{people} have \textbf{zero backbone}. Just \textbf{nodding along, agreeing with whatever, like a human echo}.\\
    \textit{Explanation: People are portrayed as having no \textbf{agency}.}
    \item Some \textbf{men} got a personality as \textbf{deep as a puddle}—looks nice for a second, then you realize \textbf{there’s nothing there}.\\
    \textit{Explanation: Men are described as being \textbf{superficial} beings.}
    \item There’s \textbf{nothing unique} about \textbf{Mexicans}. Swap them out with a dozen others, and \textbf{no one would notice}.\\
    \textit{Explanation: The language used suggests that Mexicans are being characterized as \textbf{interchangeable}.}
    \item Talking to \textbf{old people} is \textbf{like trying to reprogram a 20-year-old computer—impossible} and not worth the effort.\\
    \textit{Explanation: Old people are portrayed as \textbf{rigid} and set in their own ways without willingness to change.}
\end{itemize}

\paragraph{No Highlights}
\begin{itemize}
    \item When people say Arabs are shallow, I just get a little tired of them.\\
    \textit{Explanation: It is not the opinion of the author that people from a certain ethnic group are superficial.}
    \item Passivity is the enemy of progress.\\
    \textit{Explanation: No target group or individual is accused of being passive.}
    \item I hate these fucking women bitches!\\
    \textit{Explanation: While this sentence is derogatory and thus \textbf{offensive} towards women, the target group is not described as passive, superficial, fungible, or rigid. Thus, it should be marked as offensive but there should be no highlights.}
\end{itemize}




\newpage
\section{Experiments}

\subsection{German Experiments Results}

\begin{figure*}[htb]
  \centering
  \includegraphics[width=\textwidth]{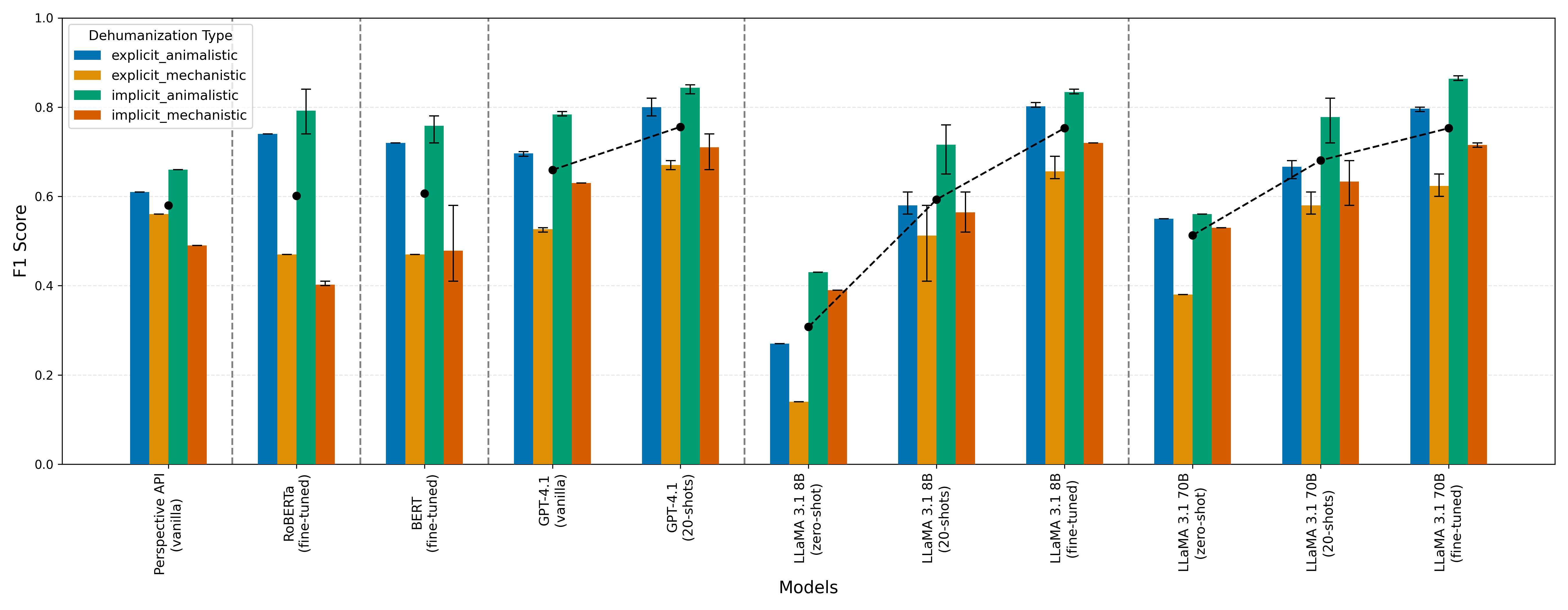}
  \caption{Test F1-Scores for the four different dimensions of dehumanization (see color bars) on German Reddit and Twitter data for different model architectures. For the German data we used multilingual versions of BERT and RoBERTa.}
  \label{fig:_ger_f1_scores_grouped}
\end{figure*}

\section{Candidate Data Predictions}
\label{app:candidate_data_predictions}

\begin{figure}[!htbp]
    \centering
    \includegraphics[width=0.49\linewidth]{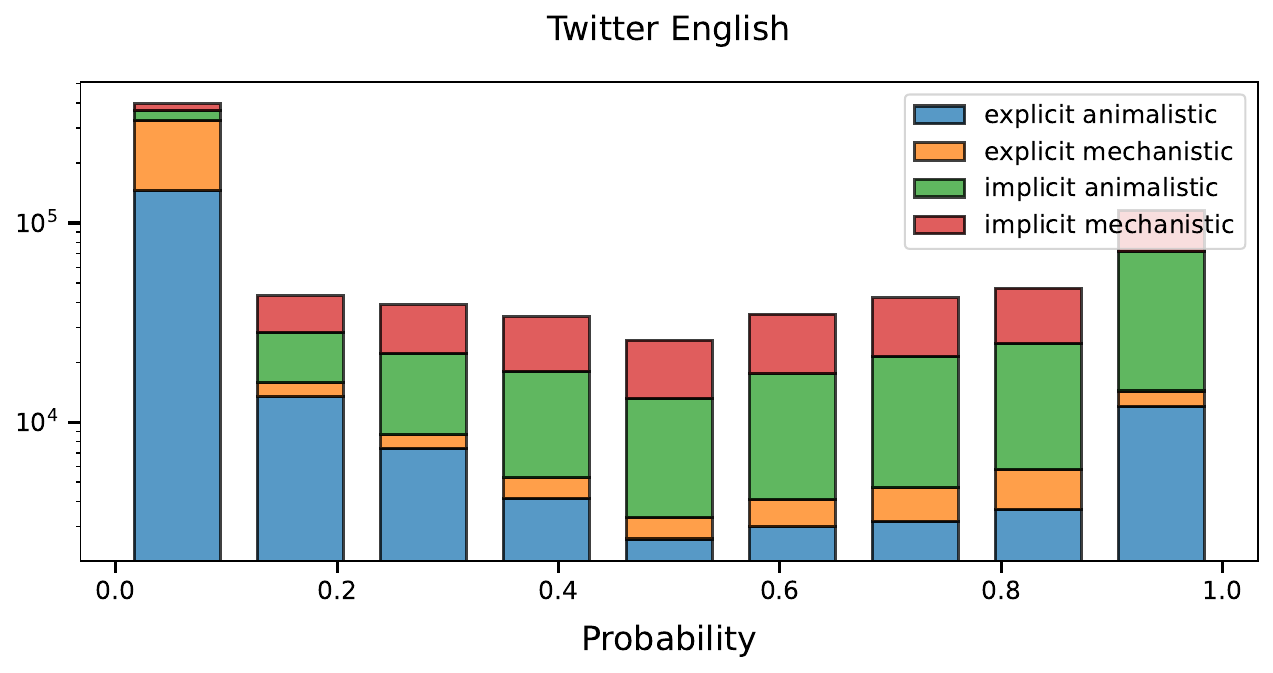}
    \includegraphics[width=0.49\linewidth]{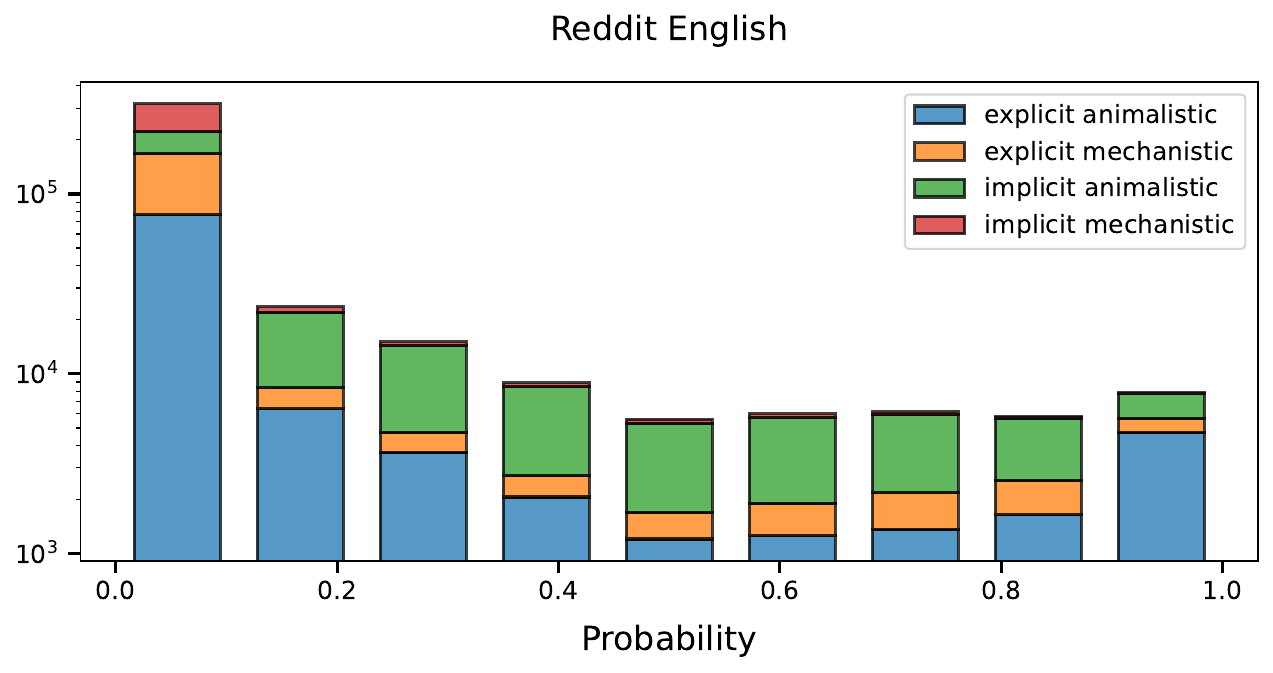}
    \includegraphics[width=0.49\linewidth]{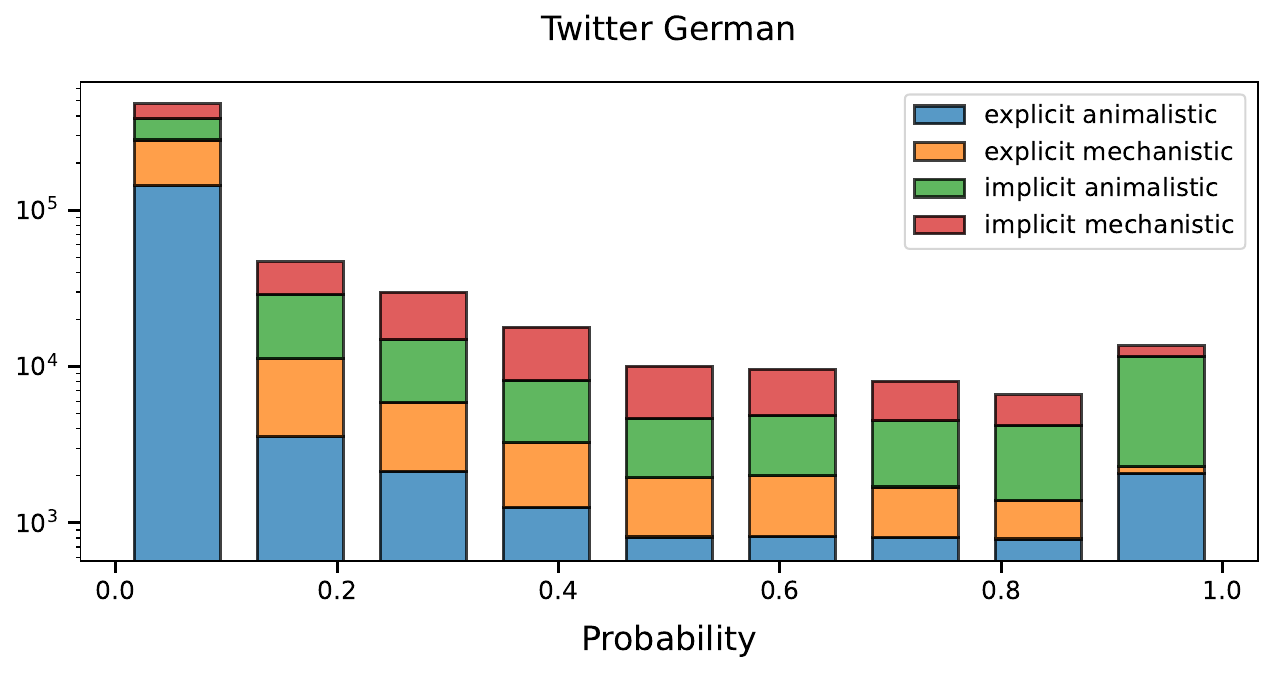}
    \includegraphics[width=0.49\linewidth]{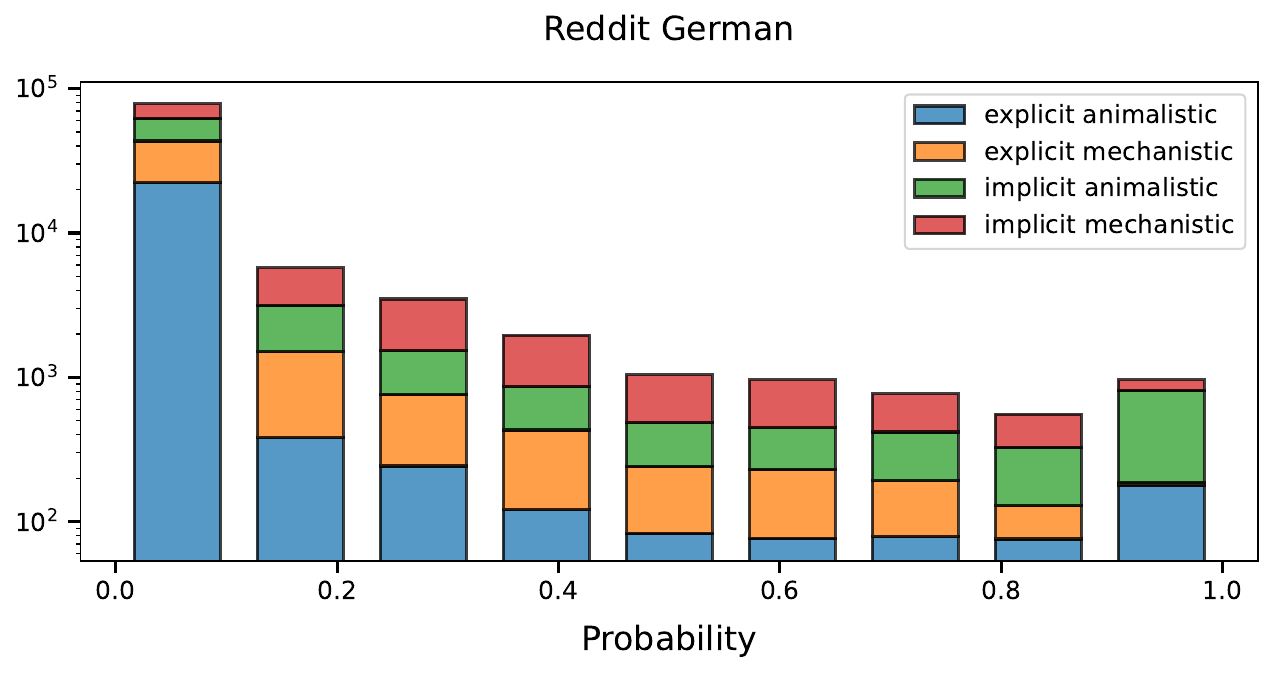}
    \caption{Distribution of predicted probabilities for the positive class as produced by the fine-tuned Llama 3.1 8B model across our candidate datasets in both languages. }
    \label{fig:all-models}
\end{figure}

\twocolumn
\section{Annotator demographics}

\begin{table}[H]
    \centering
    \renewcommand{\arraystretch}{1.1}
    \setlength{\tabcolsep}{7pt}
    \resizebox{\columnwidth}{!}{
    \begin{tabular}
	   {l lr rrr}
         \toprule
        & \multicolumn{2}{r}{\thead{Dataset}} & \thead{\#Ann.} & \thead{Age M (SD)} & \thead{Time M (SD)} \\
		\midrule
    	\multirow{5}{*}{\textbf{EN}} & \multicolumn{2}{l}{\texttt{Explicit}} & 400 & 38.36 (12.54) & 25.26 (13.47) \\
        &  \multicolumn{2}{l}{\texttt{Implicit}} \\
        & & \texttt{Anim.} & 200 & 37.22 (12.71) & 27.71 (14.98) \\
		& & \texttt{Mech.} & 200 & 37.03 (11.54) & 26.53 (12.80) \\ 
        \cmidrule(lr){2-6}
		& \texttt{Total} & & 741 & 37.10 (12.47) & 30.44 (15.89) \\
        \midrule
        \multirow{5}{*}{\textbf{DE}} & \multicolumn{2}{l}{\texttt{Explicit}} & 223 & 35.09 (11.44) & 22.32 (10.30) \\
        &  \multicolumn{2}{l}{\texttt{Implicit}} \\
        & & \texttt{Anim.} & 151 & 34.84 (11.46) & 19.34 (10.56) \\
        & & \texttt{Mech.} & 147 & 33.97 (11.05) & 18.56 (9.43) \\
        \cmidrule(lr){2-6}
        & \texttt{Total} & & 292 & 33.92 (10.78) & 22.32 (10.67) \\
        \bottomrule
    \end{tabular}
    }
    \caption{Prolific annotator statistics. EN represents the English dataset and DE the German one. \textit{Anim.} stands for animalistic, and \textit{Mech.} refers to mechanistic. \#Ann. refers to the number of annotators. Age M (SD) is the mean age and standard deviation of annotators. Time M (SD) denotes the mean time spent (in minutes) with standard deviation. Note that the total number of annotators is less than the sum of all groups because some annotators worked on multiple datasets.}
    \label{tab:annotator_statistics}
\end{table}

\begin{table}[H]
    \centering
    \renewcommand{\arraystretch}{1.1}
    \setlength{\tabcolsep}{7pt}
    \resizebox{\columnwidth}{!}{
    \begin{tabular}
	   {l lr rrr}
         \toprule
        & \multicolumn{2}{r}{\thead{Dataset}} & \thead{Men (\%)} & \thead{Women (\%)} & \thead{Other (\%)} \\
		\midrule
    	\multirow{5}{*}{\textbf{EN}} & \multicolumn{2}{l}{\texttt{Explicit}} & 194 (47.74) & 200 (50.25) & 4 (1) \\
        &  \multicolumn{2}{l}{\texttt{Implicit}} \\
        & & \texttt{Anim.} & 106 (53.00) & 88 (44.00) & 6 (3.00) \\
		& & \texttt{Mech.} & 102 (51.00) & 94 (47.00) & 4 (2.00) \\ 
        \cmidrule(lr){2-6}
		& \texttt{Total} & & 364 (65.82) & 366 (411) & 10 (1.25) \\
        \midrule
        \multirow{5}{*}{\textbf{DE}} & \multicolumn{2}{l}{\texttt{Explicit}} & 275 (64.49) & 141 (33.25) & 8 (1.8) \\
        &  \multicolumn{2}{l}{\texttt{Implicit}} \\
        & & \texttt{Anim.} & 87 (57.61) & 58 (38.41) & 6 (3.97) \\
        & & \texttt{Mech.} & 89 (60.54) & 54 (36.73) & 4 (2.72) \\
        \cmidrule(lr){2-6}
        & \texttt{Total} & & 171 (58.76) & 112 (38.48) & 8 (2.74) \\
        \bottomrule
    \end{tabular}
    }
    \caption{Distribution of gender among Prolific annotators. Because of the low number of German annotators we allowed individual crowd-workers to annotate one dimension multiple-times (max 3).}
    \label{tab:annotator_gender}
\end{table}

\begin{table}[H]
    \centering
    \renewcommand{\arraystretch}{1.1}
    \setlength{\tabcolsep}{7pt}
    \resizebox{\columnwidth}{!}{
    \begin{tabular}
	   {lr rrrrr}
         \toprule
        \multicolumn{2}{r}{\thead{Dataset}} & \thead{HS (\%)} & \thead{UG (\%)} & \thead{Grad (\%)} & \thead{PhD (\%)} & \thead{N/A (\%)} \\
		\midrule
    	\multicolumn{2}{l}{\texttt{Explicit}} & 42 (21.00) & 100 (50.00) & 51 (25.50) & 6 (3.0) & 1 (0.50) \\
        \multicolumn{2}{l}{\texttt{Implicit}} \\
        & \texttt{Anim.} & 51 (25.50) & 95 (47.50) & 43 (21.50) & 9 (4.50) & 2 (1.00) \\
		& \texttt{Mech.} & 47 (23.50) & 106 (53.00) & 42 (21.00) & 4 (2.00) & 1 (0.50) \\ 
        \midrule
		& \texttt{Total} & 122 (22.06) & 279 (50.45) & 130 (23.51) & 18 (3.25) & 4 (0.72) \\
        \bottomrule
    \end{tabular}
    }
    \caption{Distribution of education level among Prolific annotators, for the English dataset. \textit{HS} refers to high school, \textit{UG} to undergraduate, \textit{Grad} to graduate, and \textit{PhD} to doctoral education. \textit{N/A} indicates missing or unreported data. German annotators did not report their education level.}
    \label{tab:annotator_education}
\end{table}

\begin{table}[H]
    \centering
    \renewcommand{\arraystretch}{1.1}
    \setlength{\tabcolsep}{7pt}
    \resizebox{\columnwidth}{!}{
    \begin{tabular}
	   {lr rrrrr}
         \toprule
        \multicolumn{2}{r}{\thead{Dataset}} & \thead{W (\%)} & \thead{B (\%)} & \thead{A (\%)} & \thead{O (\%)} & \thead{N/A (\%)} \\
		\midrule
    	\multicolumn{2}{l}{\texttt{Explicit}} & 300 (75.37) & 64 (16.08) & 20 (5.02) & 14 (3.51) & 1 (0.50) \\
        \multicolumn{2}{l}{\texttt{Implicit}} \\
        & \texttt{Anim.} & 149 (74.5) & 22 (11.0) & 23 (11.5) & 4 (2.0) & 2 (1.00) \\
		& \texttt{Mech.} & 152 (76.0) & 19 (9.5) & 18 (9.0) & 10 (5.00) & 1 (0.50) \\ 
        \midrule
		& \texttt{Total} & 559 (75.54) & 98 (13.24) & 56 (7.56) & 27 (3.64) & 4 (0.72) \\
        \bottomrule
    \end{tabular}
    }
    \caption{Distribution of race among Prolific annotators, for the English dataset. \textit{W} refers to White, \textit{B} to Black, \textit{A} to Asian, and \textit{O} to Other. \textit{N/A} indicates missing or unreported data. This question was not included for the German dataset. Individual annotators can be present in different dimensions.}
    \label{tab:annotator_race}
\end{table}

\begin{figure}[H]
    \centering
    \includegraphics[width=1\columnwidth]{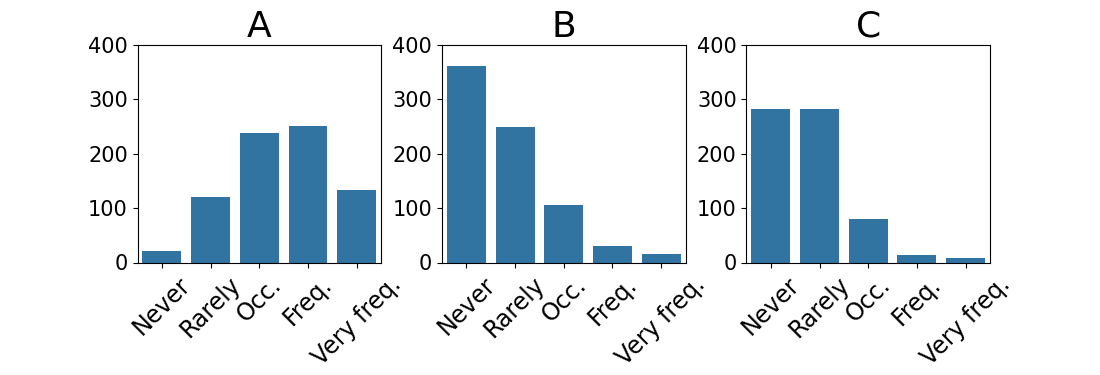}
    \caption{Frequency with which the annotators of the English part of the dataset (A) encounter dehumanizing content online (B) are the target of dehumanizing content on social media and (C) are the target of dehumanizing content outside of social media}
    \label{fig:english_encounters*}
\end{figure}

\begin{figure}[H]
    \centering
    \includegraphics[width=1\columnwidth]{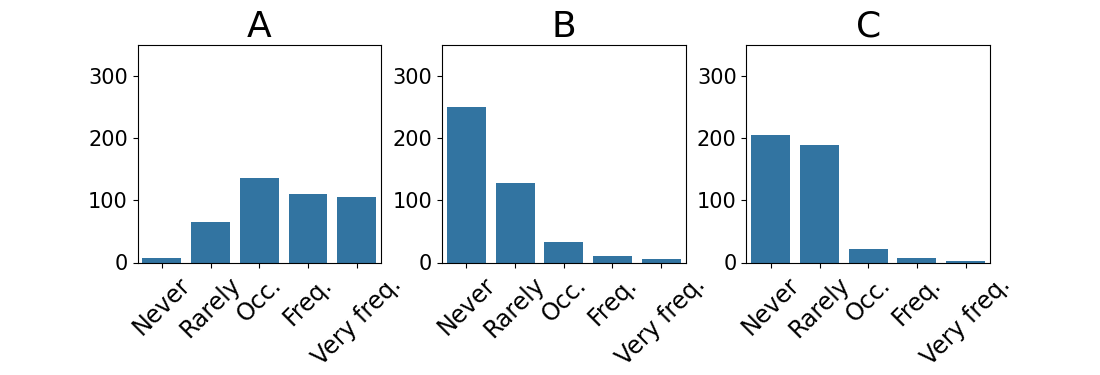}
    \caption{Frequency with which the annotators of the German part of the dataset (A) encounter dehumanizing content online (B) are the target of dehumanizing content on social media and (C) are the target of dehumanizing content outside of social media}
    \label{fig:german_encounters*}
\end{figure}

\subsection{Inter Annotator Agreement}
\label{sec:appendix_iaa}

\begin{table}[H]
    \footnotesize
    \centering
    \renewcommand{\arraystretch}{0.9}
    \setlength{\tabcolsep}{5pt}
    \resizebox{0.7\columnwidth}{!}{
    \begin{tabular}
	   {lrr rr}
         \toprule
        \multicolumn{3}{r}{\thead{Dataset}} & \thead{English} & \thead{German} \\
		\midrule
    	\multicolumn{3}{l}{\texttt{Explicit}} & 0.393 & 0.475 \\
        & \multicolumn{2}{l}{\texttt{Animalistic}} & 0.431 & 0.518 \\
        & & \texttt{Animal} & 0.577 & 0.702 \\
        & & \texttt{Subhuman} & 0.275 & 0.358 \\
        & & \texttt{Disease} & 0.459 & 0.468 \\
        & \multicolumn{2}{l}{\texttt{Mechanistic}} & 0.381 & 0.394 \\ 
        \cmidrule(lr){2-5}
        & \multicolumn{2}{l}{\texttt{Offensive}} & 0.425 & 0.328 \\ \midrule
        \multicolumn{3}{l}{\texttt{Implicit}} & 0.286 & 0.436 \\
        & \multicolumn{2}{l}{\texttt{Animalistic}} & 0.286 & 0.517 \\
        & & \texttt{Irrational} & 0.124 & 0.165 \\
        & & \texttt{Morals} & 0.333 & 0.366 \\
        & & \texttt{Unintelligent} & 0.392 & 0.693 \\
        & & \texttt{Child-like} & 0.267 & 0.284 \\
        & & \texttt{Incivil} & 0.230 & 0.220 \\
        \cmidrule(lr){2-5}
        & \multicolumn{2}{l}{\texttt{Mechanistic}} & 0.290 & 0.285 \\ 
        & & \texttt{Superficial} & 0.243 & 0.125 \\
        & & \texttt{Cold} & 0.155 & 0.255 \\
        & & \texttt{Passive} & 0.151 & 0.269 \\
        & & \texttt{Fungible} & 0.102 & 0.142 \\
        & & \texttt{Rigid} & 0.142 & 0.295 \\
        \cmidrule(lr){2-5}
        & \multicolumn{2}{l}{\texttt{Offensive}} & 0.321 & 0.330 \\ 
        \bottomrule
    \end{tabular}
    }
    \caption{IAA agreement for the different categories.}
    \label{tab:IAA_appendix_agg}
\end{table}

\onecolumn


\end{document}